\documentclass[twocolumn,10pt]{jmlr} 

\usepackage{booktabs}
\usepackage{siunitx}
\usepackage{xcolor}
\usepackage{enumitem}
\usepackage{makecell}
\usepackage{amsmath}
\usepackage{arydshln}
\usepackage{tabularray}
\usepackage{enumitem}
\usepackage{stfloats}
\usepackage[most]{tcolorbox}
\usepackage{amsmath,amssymb}
\usepackage{todonotes}
\colorlet{darkgreen}{green!65!black}
\usepackage[export]{adjustbox}%
\usepackage{lscape}
\usepackage{placeins}
\usepackage{float}

\usepackage{tikz}
\usepackage{multirow}
\usepackage{array}
\usepackage{booktabs}
\usepackage{graphicx}
\usepackage{soul}
\definecolor{Gray}{gray}{0.85}

\definecolor{GREEN}{rgb}{0.21, 0.63, 0.18}  
\definecolor{RED}{rgb}{0.89, 0.09, 0.11}  

\definecolor{uc_color}{rgb}{0.99,0.24,0.63}
\definecolor{hc_color}{rgb}{0.02,0.51,0.51}
\definecolor{tc_color}{rgb}{0.99,0.55,0.09}

\def\boxit#1#2{%
    \smash{\color{black}\fboxrule=0.5pt\relax\fboxsep=2.5pt\relax%
    \llap{\rlap{\fbox{\phantom{\rule{#1}{#2}}}}~}}\ignorespaces
}

\usepackage[switch]{lineno}

\theorembodyfont{\upshape}
\theoremheaderfont{\scshape}
\theorempostheader{:}
\theoremsep{\newline}

\jmlrvolume{} 
\jmlryear{2024} 
\jmlrsubmitted{} 
\jmlrpublished{} 
\jmlrworkshop{Conference on Health, Inference, and Learning (CHIL) 2024} 

\title[Health-LLM: Large Language Models for Health Prediction via Wearable Sensor Data]{Health-LLM: Large Language Models for Health Prediction via Wearable Sensor Data}

\author{%
\Name{Yubin Kim} \Email{ybkim95@mit.edu}\\
\addr MIT, Cambridge, MA
\AND
\Name{Xuhai Xu} \Email{xoxu@mit.edu}\\
\addr MIT, Cambridge, MA
\AND
\Name{Daniel McDuff} \Email{dmcduff@google.com}\\
\addr Google Research, WA
\AND
\Name{Cynthia Breazeal} \Email{cynthiab@mit.edu}\\
\addr MIT, Cambridge, MA
\AND
\Name{Hae Won Park} \Email{haewon@mit.edu}\\
\addr MIT, Cambridge, MA
}

\begin{document}

\maketitle

\begin{abstract}
Large language models (LLMs) are capable of many natural language tasks, yet they are far from perfect. In health applications, grounding and interpreting domain-specific and non-linguistic data is crucial. This paper investigates the capacity of LLMs to make inferences about health based on contextual information (e.g. user demographics, health knowledge) and physiological data (e.g. resting heart rate, sleep minutes). We present a comprehensive evaluation of 12 state-of-the-art LLMs with prompting and fine-tuning techniques on four public health datasets (PMData, LifeSnaps, GLOBEM and AW\_FB). Our experiments cover 10 consumer health prediction tasks in mental health, activity, metabolic, and sleep assessment. Our fine-tuned model, HealthAlpaca exhibits comparable performance to much larger models (GPT-3.5, GPT-4 and Gemini-Pro), achieving the best performance in \textbf{8 out of 10} tasks. Ablation studies highlight the effectiveness of context enhancement strategies. Notably, we observe that our context enhancement can yield up to \textbf{23.8\%} improvement in performance. While constructing contextually rich prompts (combining user context, health knowledge and temporal information) exhibits synergistic improvement, the inclusion of health knowledge context in prompts significantly enhances overall performance.
\end{abstract}

\paragraph*{Data and Code Availability}
In this study, we conduct experiments with 4 public datasets: \\ (1) PMData \footnote{\url{https://datasets.simula.no/pmdata/}} (\cite{10.1145/3339825.3394926}), (2) LifeSnaps \footnote{\url{https://github.com/Datalab-AUTH/LifeSnaps-EDA}}(\cite{yfantidou2022lifesnaps}), (3) GLOBEM \footnote{\url{https://the-globem.github.io/datasets/overview}} (\cite{xu2022globem}) and (4) AW\_FB \footnote{\url{https://dataverse.harvard.edu/dataset.xhtml?persistentId=doi:10.7910/DVN/ZS2Z2J}} (\cite{DVN/ZS2Z2J_2020}). Code is available at \url{https://github.com/mitmedialab/Health-LLM}


\paragraph*{Institutional Review Board (IRB)}
This study has no human-subject research and only uses publicly available data.

\section{Introduction}
\label{sec:intro}

The performance of large language models (LLMs) \cite{openai2023gpt4, singhal2022large, Nyberg_2021} in diverse text generation and knowledge retrieval tasks presents wide-ranging opportunities~\cite{nori2023capabilities, hegselmann2023tabllm, gandhi2023understanding, wu2023bloomberggpt}. 
However, in sensitive domains like healthcare, their true capabilities and limitations remain largely unexplored, especially when it comes to harnessing the diverse collection of \emph{multi-modal}, \emph{time-series} data generated by wearable sensors. Unlike static text, this data presents unique challenges for LLMs due to its high dimensionality, non-linear relationships, and continuous nature, requiring them to understand not only individual data points but also their dynamic patterns over time. Although specialized medical-domain LLMs have shown promise in capturing domain knowledge \cite{he2023survey, singhal2022large, singhal2023towards, han2023medalpaca, toma2023clinical, thirunavukarasu2023large, mcduff2023towards}, their application to consumer health tasks, which rely heavily on physiological (e.g. heart rate) and behavioral time-series data (e.g. daily steps), remains largely untested due to the challenges of grounding LLMs in non-linguistic data and the lack of standardized evaluation benchmarks.

In this paper, we propose \textbf{Health-LLM}, a framework tailored to the healthcare domain that aims to bridge the gap between pre-trained knowledge in current LLMs and consumer health problems. We conducted a comprehensive evaluation of 12 state-of-the-art LLMs that are publicly accessible: MedAlpaca \citep{han2023medalpaca}, PMC-Llama \citep{wu2023pmc}, Llama 2 \citep{touvron2023llama}, BioMedGPT\citep{zhang2024biomedgpt}, BioMistral \citep{labrak2024biomistral}, Asclepius \citep{kweon2023publicly}, ClinicalCamel \citep{toma2023clinical}, Flan-T5 \citep{chung2022scaling}, Palmyra-Med \citep{unknown}, GPT-3.5 \citep{gpt-3.5}, GPT-4 \citep{openai2023gpt4} and Gemini-Pro \citep{geminiteam2023gemini}. We cover 10 health prediction tasks across mental health, activity tracking, metabolism and sleep assessment. Our experiments include four steps: (i) zero-shot prompting, (ii) few-shot prompting along with chain-of-thoughts (CoT) and self-consistency (SC) prompting, (iii) instructional fine-tuning and (iv) ablation studies with context enhancement in \textit{zero-shot} setting, where context enhancement refers to the strategic inclusion of additional information - 1) user profile, 2) health knowledge, 3) temporal context, and 4) a combination of these in the prompts for LLMs to improve their understanding in healthcare domain.  

We found in (i) that zero-shot prompting shows comparable results to task-specific baseline models.
Comparing (i) and (ii) shows that few-shot prompting with bigger LLMs, like GPT-3.5, GPT-4 and Gemini-Pro, can effectively ground numerical time-series data, resulting in significant improvements over zero-shot learning and fine-tuned models in some tasks.
Through step (iii), our Alpaca-based fine-tuned model, namely \textit{HealthAlpaca}, exhibits the best performance in 8 out of 10 tasks despite being a substantially smaller than GPT-3.5, GPT-4 and Gemini-Pro.
In (iv), our ablation study indicates that the context enhancement strategy yields up to 23.8\% performance improvement, emphasizing the importance of contextual information in the prompt for LLMs in the healthcare domain. Finally, we present two case studies demonstrating the step-by-step reasoning process of representative LLMs in health prediction tasks, illustrating their ability to capture time-series data and offer personalized recommendations.

The contribution of our paper can be summarized as follows:

\begin{figure*}[t] 
    \centering
    \includegraphics[width=1.0\linewidth]{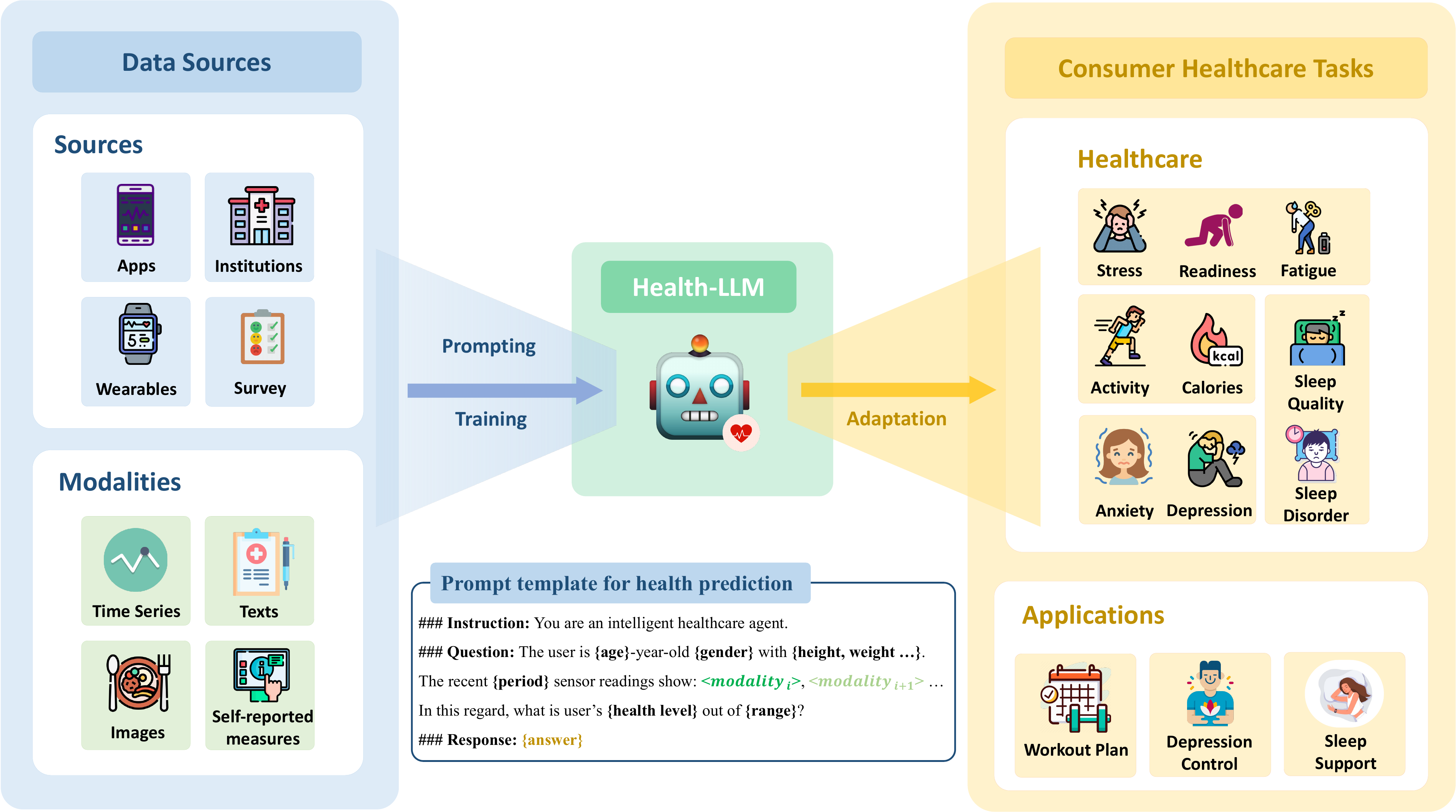}
    \caption{\textbf{Health-LLM.} We present a framework for evaluating LLM performance on a diverse set of health prediction tasks, training and prompting the models with multi-modal health data.}
    \label{fig:framework}
\end{figure*}

\begin{itemize}
  \item We present Health-LLM, a framework that enables LLMs to adapt to health predictions by prompting/training via wearable sensor data.
  \item We combine four publicly available health datasets, curate ten novel consumer health tasks, and conduct evaluations with twelve state-of-the-art LLMs.
  \item We show the effectiveness of context enhancement strategies for Health-LLMs and release our fine-tuned model HealthAlpaca, as the first set of open-source LLMs targeted for consumer health prediction tasks.
  
\end{itemize}

\section{Related Work}

\subsection{Wearable Sensor Data with LLMs}
Wearable sensor technology has transformed personal health monitoring, enabling continuous tracking of vital physiological data such as heart rate variability and step counts \cite{doi:10.1056/NEJMoa1901183}. These time-series data can be represented in different forms, including statistical summaries (e.g., mean, standard deviation), Fourier transforms, and more.

The representation of data before feeding it into the model can also vary. Data can be input as raw time-series text, where the textual information is directly used, or as encoder embeddings obtained from modality-specific \citep{belyaeva2023multimodal} and timestamp encoders \citep{zhou2022fedformer, zhou2021informer}. Also, the integration of temporal information has proven to be effective in enhancing outcomes \citep{wen2023transformers}.

The synergy between wearable sensor data and advanced machine learning techniques holds promise in predicting diverse health outcomes, such as depression scores \citep{info:doi/10.2196/35807,englhardt2023classification}, early detection of atrial fibrillation \citep{chen2022atrial}, and monitoring stress levels \citep{VOS2023104556}. Additionally, this integration facilitates personalized monitoring in areas like nutrition \citep{sempionatto2021wearable, romero2023ai4fooddb} and stress management \citep{tazarv2021personalized}.

\subsection{Health LLMs}
The integration of LLMs in healthcare is a rapidly growing research field \citep{tu2024conversational, wang2023drgllama, liu2023large, han2023medalpaca, tang2023medagents, belyaeva2023multimodal}. For instance, \cite{singhal2022large} demonstrated the efficacy of LLMs through a combination of base improvements (PaLM 2), medical domain fine-tuning, and ensemble refinement, outperforming benchmarks across various datasets. Additionally, \cite{xu2023mentalllm} explored LLM adaptation for diverse mental health tasks using online-text data, employing techniques like zero-shot, few-shot, and fine-tuning, while our work covers multi-modal time-series data collected from wearable sensors. \\
In a recent comprehensive evaluation of GPT-4 \citep{nori2023capabilities}, the general-purpose model without fine-tuning, surpassed the USMLE passing score by over 20 points, outperforming earlier models like GPT-3.5 and medically fine-tuned models. This study showed GPT-4's qualitative abilities in explaining medical reasoning, personalizing explanations, and crafting counterfactual scenarios. 

\begin{table*}[t!]
\tiny
\caption{\textbf{Different Types of Contexts in Health Prompts}.}
\begin{tabular}{>{\centering\arraybackslash}p{2.6cm} >{\centering\arraybackslash}p{13.0cm}}
\toprule
\textbf{Context} & \textbf{Prompt} \\
\midrule
\multirow{3}{*}{Basic ($bs$)} & \begin{minipage}[t]{\linewidth}
The analysis of recent period: \{14\} days averaged sensor readings show: Steps: \{812.0\} steps, Burned Calories: \{97.0\} calories, Resting Heart Rate: \{66.54\} beats/min, Sleep Minutes: \{487.0\} minutes, Mood: \{3\} out of 5. In this regard, what is the predicted readiness score/level between 0 and 10?
\end{minipage}\\ 
\midrule
\multirow{4}{*}{User Context ($uc$)} & \begin{minipage}[t]{\linewidth}
\tcbox[on line, boxsep=1pt, left=0pt,right=0pt,top=0pt,bottom=0pt,colframe=uc_color!30, colback=uc_color!30]{Given the user's profile as age: \{23\}-year-old, sex: \{male\} and height: \{182\} cm}, the analysis of recent period: \{14\} days averaged sensor readings show: Steps: \{812.0\} steps, Burned Calories: \{97.0\} calories, Resting Heart Rate: \{66.54\} beats/min, Sleep Minutes: \{487.0\} minutes, Mood: \{3\} out of 5. In this regard, what is the predicted readiness score/level between 0 and 10?
\end{minipage}\\  
\midrule
\multirow{4}{*}{Health Context ($hc$)} & \begin{minipage}[t]{\linewidth}
\tcbox[on line, boxsep=1pt, left=0pt,right=0pt,top=0pt,bottom=0pt,colframe=hc_color!30, colback=hc_color!30]{Readiness score is an indicator of how prepared our body is for physical activity. It is decided by activity, recent sleep} \tcbox[on line, boxsep=1pt, left=0pt,right=0pt,top=0pt,bottom=0pt,colframe=hc_color!30,colback=hc_color!30]{and heart rate variability.} The analysis of recent period: \{14\} days averaged sensor readings show: Steps: \{812.0\} steps, Burned Calories: \{97.0\} calories, Resting Heart Rate: \{66.54\} beats/min, Sleep Minutes: \{487.0\} minutes, Mood: \{3\} out of 5. In this regard, what is the predicted readiness score/level between 0 and 10?
\end{minipage}\\ 
\midrule
\multirow{5}{*}{Temporal Context ($tc$)} & \begin{minipage}[t]{\linewidth}
The analysis of recent period: \{14\} days sensor readings show: Steps: \tcbox[on line, boxsep=1pt, left=0pt,right=0pt,top=0pt,bottom=0pt,colframe=tc_color!30, colback=tc_color!30]{\{"NaN, 991.0, ..., NaN"\}} steps, Burned Calories: \tcbox[on line, boxsep=1pt, left=0pt,right=0pt,top=0pt,bottom=0pt,colframe=tc_color!30, colback=tc_color!30]{\{"NaN, 94.0 ..., NaN"\}} calories, Resting Heart Rate: \tcbox[on line, boxsep=1pt, left=0pt,right=0pt,top=0pt,bottom=0pt,colframe=tc_color!30, colback=tc_color!30]{\{"69.32, 67.72, ..., 64.55"\}} beats/min, Sleep Minutes: \tcbox[on line, boxsep=1pt, left=0pt,right=0pt,top=0pt,bottom=0pt,colframe=tc_color!30, colback=tc_color!30]{\{"534.0, 455.0, ..., 405.0"\}} minutes, [Mood]: 3 out of 5. In this regard, what is the predicted readiness score/level between 0 and 10?
\end{minipage}\\  
\midrule
\multirow{8}{*}{All ($all$)} & \begin{minipage}[t]{\linewidth}
\tcbox[on line, boxsep=1pt, left=0pt,right=0pt,top=0pt,bottom=0pt,colframe=hc_color!30, colback=hc_color!30]{Readiness score is an indicator of how prepared our body is for physical activity. It is decided by activity, recent sleep} \tcbox[on line, boxsep=1pt, left=0pt,right=0pt,top=0pt,bottom=0pt,colframe=hc_color!30, colback=hc_color!30]{and heart rate variability.} \tcbox[on line, boxsep=1pt, left=0pt,right=0pt,top=0pt,bottom=0pt,colframe=uc_color!30, colback=uc_color!30]{The user is 23-year-old male with 182 cm.} The analysis of recent period: \{14\} days sensor readings show: Steps: \tcbox[on line, boxsep=1pt, left=0pt,right=0pt,top=0pt,bottom=0pt,colframe=tc_color!30, colback=tc_color!30]{\{"NaN, 991.0, ..., NaN"\}} steps, Burned Calories: \tcbox[on line, boxsep=1pt, left=0pt,right=0pt,top=0pt,bottom=0pt,colframe=tc_color!30, colback=tc_color!30]{\{"NaN, 94.0 ..., NaN"\}} calories, Resting Heart Rate: \tcbox[on line, boxsep=1pt, left=0pt,right=0pt,top=0pt,bottom=0pt,colframe=tc_color!30, colback=tc_color!30]{\{"69.32, 67.72, ..., 64.55"\}} beats/min, Sleep Minutes: \tcbox[on line, boxsep=1pt, left=0pt,right=0pt,top=0pt,bottom=0pt,colframe=tc_color!30, colback=tc_color!30]{\{"534.0, 455.0, ..., 405.0"\}} minutes, [Mood]: 3 out of 5. In this regard, what is the predicted readiness score/level between 0 and 10?
\end{minipage}\\  
\bottomrule
\label{tab:context_enhancement}
\end{tabular}
\end{table*}

\begin{figure*}
\begin{align}
\text{Prompt}_{ZS} &= \text{Instruction}_{ZS} \ + \ \underline{\text{Context}} \ + \ \text{Question} \ + \ \text{Output Format}  \label{eq:1} \\
\underline{\text{Context}} &= C_{health}^{ *} \ + \ C_{user}^{ *} \ + \ TE^{*}(\text{TimeSeriesData}) \label{eq:2} \\
\text{Prompt}_{FS} &= \text{Instruction}_{FS} \ + \ (\underline{\text{Context}} + \text{Question}, \ \text{Answer})_{N} \ + \ \text{Prompt}_{ZS}  
\label{eq:3}
\end{align}
\end{figure*}

\section{Methods}
\label{sec:math}

\subsection{Zero-shot Prompting}
The aim of zero-shot prompting is to evaluate the capability of pre-trained knowledge in LLMs on health prediction tasks. To this end, we first design a basic prompt setup (\textit{bs}) that summarizes the wearable sensor data into a paragraph. Then, we introduce a comprehensive zero-shot prompting along with four types of \textit{context enhancements} introduced in Table \ref{tab:context_enhancement} and Equation \ref{eq:1}-\ref{eq:2}. 1) \textbf{User Context} ($uc$) provides user-specific information such as age, gender, weight, height, etc., which provides additional information that affects the understanding of health knowledge. 2) \textbf{Health Context} ($hc$) provides the definition and equation that controls certain health targets to inject new health knowledge into LLMs. 3) \textbf{Temporal Context} ($tc$) is adopted to test the importance of temporal aspects in time-series data. Instead of using aggregated statistics, we utilize the raw time-series sequence. Among different sets of temporal context representations, we empirically observe that using natural language string showed the best performance. 
In Section \ref{sec:temporal_encoding}, we introduce three methods for Temporal Encoder $TE(\cdot)$.  

\subsection{Few-shot Prompting}

Few-shot prompting involves using a limited selection of demonstration examples within the prompts to facilitate in-context learning.
In our case, we adopt 3-shot setting. These demonstrations are used exclusively within the prompts, while the model parameters remain static. This approach is akin to providing the model with a handful of case studies to help it grasp and apply healthcare domain knowledge effectively. In addition to few-shot prompting, we further enriched the prompting strategy by integrating Chain-of-Thoughts (CoT) \citep{wei2022chain} and Self-Consistency (SC) \citep{wang2022self} prompting techniques. The incorporation of CoT prompting facilitates a more cohesive and contextually nuanced understanding, allowing the model to connect ideas seamlessly. Simultaneously, SC prompting contributed to refining the model's responses by promoting internal coherence and logical consistency. Together, these methodologies synergistically formed a robust prompt, leveraging diverse prompting techniques to optimize the LLMs health understanding capabilities \citep{wang2023selfconsistency}. In this work, we used $N=5$ candidate reasoning paths for SC prompting.

\subsection{Instruction Tuning}
Instruction tuning is a technique where all parameters of a pre-trained model are further trained or fine-tuned on a target task. This process allows the model to adapt its pre-trained knowledge to the specificities of the new task, optimizing its performance. 
In the context of health prediction, fine-tuning allows the model to deeply understand physiological terminologies, mechanisms, and context, thereby enhancing its ability to generate accurate and contextually relevant responses.

Instead of fine-tuning all parameters, methods like LoRA \citep{hu2021lora} involves training a small proportion of parameters, by injecting trainable low-rank matrices into each layer of the pre-trained model. In the Health-LLM context, these Parameter Efficient Fine-tuning (PEFT) techniques enable the model to adapt to healthcare tasks while maintaining computational efficiency. 

\vspace{-4pt}

\subsection{Temporal Encoding Methods}
The methodologies for encoding time-series data into textual formats is crucial in performing health predictions with LLMs. In the context of our consumer health prediction task, we selected to use the \textbf{Natural Language String} method. This was due to its simplicity, effectiveness, and wide acceptance in the field, as evidenced by various prior studies \citep{gruver2023large, liu2023large}. This method's inherent interpretability aligns seamlessly with our objective to develop models that are easily understood and manipulated by users. Furthermore, it's capacity to handle missing values without imputation by adopting special symbols (e.g. NaN), making them more resilient to data corruption, is highly advantageous for health-related tasks where data may often be incomplete or irregular. Lastly, our decision was further influenced by the organization of physiological data in our datasets by specific time windows (daily, weekly, monthly), which are intuitively represented through natural language. This facilitates a clearer understanding and processing by LLMs, enhancing our model's capability to make nuanced predictions. We demonstrate more methods in Figure \ref{fig:time-series_encoding1} and in Appendix \ref{apd:second}, by introducing three popular methods: 1) Natural Language String, 2) Modality-specific Encoding, and 3) Statistical Summary.
\label{sec:temporal_encoding}

\begin{table*}[t]
\tiny
\centering
\caption{\textbf{Consumer Health Tasks}. We define thirteen tasks from six datasets and classify them into four topics. * in the prompt indicates the optional contexts for the ablation.}
\begin{tabular}{>{\rotatebox{90}}p{1.1cm} >{\centering\arraybackslash}p{1.0cm} >{\centering\arraybackslash}p{2.3cm} p{1.1cm} p{6.5cm} >{\centering\arraybackslash}p{1.5cm}}
\toprule
\centering\textbf{Topic} & \centering\textbf{Dataset} & \centering\textbf{Task} & \centering\textbf{Metric} & \centering\textbf{Prompt} & \textbf{Target} \\
\midrule
\multirow{20}{*}{MHealth} & \multirow{4}{*}{PMData} & \multirow{4}{*}{Stress Prediction} & \centering \multirow{4}{*}{MAE $\downarrow$}  & \textcolor{blue}{\upshape{\{Target\}}} refers to \textcolor{teal}{\upshape{[Health Knowledge]*}}. Given the \textcolor{magenta}{\upshape{[User Info]*}}, and \textcolor{orange}{\upshape{[Period]}} sequence of Steps: \textcolor{orange}{\upshape{[Steps]}}, Calories Burn: \textcolor{orange}{\upshape{[Calories]}}, Resting Heart Rate: \textcolor{orange}{\upshape{[RHR]}}, Sleep Duration: \textcolor{orange}{\upshape{[SleepMinutes]}}, Mood: \textcolor{orange}{\upshape{[Mood]}}. What will my stress level be? & \multirow{4}{*}{\textcolor{blue}{\upshape{[Stress]}}} \\ \\
& \multirow{8}{*}{LifeSnaps} & \multirow{8}{*}{Stress Resilence Prediction} & \centering \multirow{8}{*}{MAE $\downarrow$} & \textcolor{blue}{\upshape{[Target]}} refers to \textcolor{teal}{\upshape{[Health Knowledge]*}}. Given the \textcolor{magenta}{\upshape{[User Info]*}} and following \textcolor{orange}{\upshape{[Period]}} sequence of data, predict the Stress Resilience Index. Stress Score: \textcolor{orange}{\upshape{[StressScore]}}, Positive Affect Score: \textcolor{orange}{\upshape{[PosAffectScore]}}, Negative Affect Score: \textcolor{orange}{\upshape{[NegAffectScore]}}, 
Lightly Active Minutes: \textcolor{orange}{\upshape{[Duration]}}, Moderately Active Minutes: \textcolor{orange}{\upshape{[Duration]}}, Very Active Minutes: \textcolor{orange}{\upshape{[Duration]}}, Sleep Efficiency: \textcolor{orange}{\upshape{[SleepEfficiency]}}, Sleep Deep Ratio: \textcolor{orange}{\upshape{[SleepDeepRatio]}}, Sleep Light Ratio: \textcolor{orange}{\upshape{[SleepLightRatio]}}, Sleep REM Ratio: \textcolor{orange}{\upshape{[SleepREMRatio]}}. & \multirow{7}{*}{\textcolor{blue}{\upshape{[Stress Resilience]}}} \\ \\
& \multirow{7}{*}{GLOBEM} & \multirow{7}{*}{\makecell{Estimate of PHQ4 Score}} & \centering \multirow{7}{*}{\makecell{MAE $\downarrow$ \\ MAE $\downarrow$}} & \textcolor{blue}{\upshape{[Target]}} refers to \textcolor{teal}{\upshape{[Health Knowledge]*}}. Steps during last \textcolor{orange}{\upshape{[Period]}} sequence of maximum, minimum, average, median, standard deviation daily step count were \textcolor{orange}{\upshape{[ListOfSteps]}} respectively. Sleep during last \textcolor{orange}{\upshape{[Period]}} sequence of sleep efficiency, duration the user stayed in bed after waking up, duration the user spent to fall asleep, duration the user stayed awake but still in bed, duration the user spent to fall asleep are \textcolor{orange}{\upshape{[ListOfDurations]}} in average. In this regard, what would be \textcolor{blue}{\upshape{[Target]}}?" & \multirow{7}{*}{\textcolor{blue}{\upshape{[PHQ]}}} \\
\midrule   
\multirow{7}{*}{Activity} & \multirow{4}{*}{PMData} & \multirow{4}{*}{\makecell{Readiness Prediction \\ Fatigue Prediction}} & \centering \multirow{4}{*}{\makecell{MAE $\downarrow$ \\ Accuracy $\uparrow$}} & \textcolor{blue}{\upshape{[Target]}} refers to \textcolor{teal}{\upshape{[Health Knowledge]*}}. Steps: \textcolor{orange}{\upshape{[Steps]}}, Burned Calorories: \textcolor{orange}{\upshape{[Calories]}}, Resting Heart Rate: \textcolor{orange}{\upshape{[RHR]}}, SleepMinutes: \textcolor{orange}{\upshape{[Duration]}}, Mood: \textcolor{orange}{\upshape{[Mood]}}. What will my readiness level be? & \multirow{4}{*}{\makecell{\textcolor{blue}{\upshape{[Readiness]}} \\ \textcolor{blue}{\upshape{[Fatigue]}}}} \\ \\
  & \multirow{4}{*}{AW\_FB} & \multirow{4}{*}{Activity Recognition} & \multirow{4}{*}{Accuracy $\uparrow$} & \textcolor{blue}{\upshape{[Target]}} refers to \textcolor{teal}{\upshape{[Health Knowledge]*}}. Predict the activity type among \textcolor{orange}{\upshape{[ListOfActivities]}} given the following information \textcolor{magenta}{\upshape{[User Info]*}}, Steps: \textcolor{orange}{\upshape{[Steps]}}, Burned Calorories: \textcolor{orange}{\upshape{[Calories]}}, Heart Rate: \textcolor{orange}{\upshape{[HR]}}." & \multirow{4}{*}{\textcolor{blue}{\upshape{[Activity]}}}  \\
\midrule   
\multirow{2}{*}{Metabolic} & \multirow{2}{*}{AW\_FB} & \multirow{2}{*}{Calorie Burn Estimate} & \centering \multirow{2}{*}{MAE $\downarrow$} & \textcolor{blue}{\upshape{[Target]}} refers to \textcolor{teal}{\upshape{[Health Knowledge]*}}. Predict the burned calories given the following information. \textcolor{magenta}{\upshape{[User Info]*}}, Steps: \textcolor{orange}{\upshape{[Steps]}}, Heart Rate: \textcolor{orange}{\upshape{[HR]}}. & \multirow{2}{*}{\textcolor{blue}{\upshape{[Calories]}}} \\
\midrule   
\multirow{11}{*}{Sleep} & \multirow{4}{*}{PMData} & \multirow{4}{*}{Sleep Quality Prediction} &\centering \multirow{4}{*}{MAE $\downarrow$} & \textcolor{blue}{\upshape{[Target]}} refers to \textcolor{teal}{\upshape{[Health Knowledge]*}}. Steps: \textcolor{orange}{\upshape{[Steps]}}, Burned Calorories: \textcolor{orange}{\upshape{[Calories]}}, Resting Heart Rate: \textcolor{orange}{\upshape{[RHR]}}, SleepMinutes: \textcolor{orange}{\upshape{[Duration]}}, Mood: \textcolor{orange}{\upshape{[Mood]}}. What will my sleep quality level be? & \multirow{3}{*}{ \textcolor{blue}{\upshape{[SQ]}}} \\ \\
  & \multirow{7}{*}{LifeSnaps} & \multirow{7}{*}{Sleep Disorder Prediction} & \multirow{7}{*}{Accuracy $\uparrow$} & \textcolor{blue}{\upshape{[Target]}} refers to \textcolor{teal}{\upshape{[Health Knowledge]*}}. Given the following data, predict whether there exists sleep disorder (1) or not (0). Sleep Duration: \textcolor{orange}{\upshape{[Duration]}}, Minutes Awake: \textcolor{orange}{\upshape{[Duration]}}, Sleep Efficiency: \textcolor{orange}{\upshape{[Efficiency]}}, Sleep Deep Ratio: \textcolor{orange}{\upshape{[SleepDeepRatio]}}, Sleep Wake Ratio: \textcolor{orange}{\upshape{[SleepWakeRatio]}}, Sleep Light Ratio: \textcolor{orange}{\upshape{[SleepLightRatio]}}, Sleep REM Ratio: \textcolor{orange}{\upshape{[SleepREMRatio]}}, RMSSD: \textcolor{orange}{\upshape{[RMSSD]}}, SPO2: \textcolor{orange}{\upshape{[SPO2]}}, Full Sleep Breathing Rate: \textcolor{orange}{\upshape{[BreathingRate]}}, BPM: \textcolor{orange}{\upshape{[BPM]}}, Resting Hour: \textcolor{orange}{\upshape{[Duration]}}. & \multirow{7}{*}{\textcolor{blue}{\upshape{[Sleep Disorder]}}} \\

\bottomrule
\label{tab:features_in_the_prompt}
\end{tabular}
\end{table*}

\section{Experiment}
\label{sec:vec}

\subsection{Datasets and Tasks}
We consider four wearable sensor datasets that contained: (1) multi-modal physiological data, (2) user self-reported measures, (3) enough distinct time windows to evaluate over. Table \ref{tab:consumer_health_tasks} summarizes the dataset topic, tasks, metric to evaluate, size and text length and Table \ref{tab:features_in_the_prompt} presents the features used in the prompt for each task. For the train/test split, we selected 0.1 portion of the original set as the test set and randomly sampled the data from different participants as possible. The choice of ten tasks across four datasets were inspired by the functions provided by consumer health wearables (e.g. Fitbit, Apple Watch) and the previous works of LLMs in diverse applications \citep{liu2023large, wu2023mindshift}.

\paragraph{\textbf{PMData} \cite{10.1145/3339825.3394926}} A dataset of $n=16$ participants (twelve men and three women, in the range of 25-60 years, with an average age of 34 years) during 5 months using Fitbit Versa 2 smartwatch wristbands (Objective Biometrics and Activity Data), Google Forms (Demographics, Food, Drinking, and Weight) and PMSys sports logging smartphone application (self-reported measures like fatigue, mood, stress, etc) that combines conventional lifelogging data with sports-activity data. Tasks associated with this dataset include:
\begin{itemize}[itemsep=0.05em,leftmargin=*]
\item \textbf{Stress (STRS)}: Estimation of an individual's stress level based on physiological data and self-reported measures.
\item \textbf{Readiness (READ)}: Assessment of an individual's preparedness for physical activity/exercise.
\item \textbf{Fatigue (FATG)}: Monitoring of signs of tiredness or exhaustion.
\item \textbf{Sleep Quality (SQ)}: Assessment of factors including total sleep time, efficiency, frequency and duration of awakenings during the night.
\end{itemize}

\paragraph{\textbf{LifeSnaps} \cite{yfantidou2022lifesnaps}} A comprehensive, multi-modal dataset collected over a period more than 4 months by $n=71$ participants (42 male and 29 female, half under 30 and half over 30). The annotations were collected from Fitbit Sense (Automatically Synced Data; sleep, heart rate, stress, etc) watch, SEMA3 Data (Ecological Momentary Assessments; context and mood, step goal, etc) and from the validated surveys (Self Reported Data; demographics, health, etc). Tasks associated with this dataset include:
\begin{itemize}[itemsep=0.05em,leftmargin=*]
\item \textbf{Stress Resilience (SR)}: Assessment of an individual's ability to recover from or adapt positively to stressors.
\item \textbf{Sleep Disorder (SD)}: Identification of potential sleep disorders like insomnia or sleep apnea through analysis of recorded data.
\end{itemize}

\paragraph{\textbf{GLOBEM} \cite{xu2022globem}} A multi-year passive sensing datasets over 705 user-years and $n=497$ participants' (58.9\% of females, 24.2\% of immigrants, 38.2\% of first-generations, and 9.1\% of disability, and 53.9\% of Asian and 31.9\% of White) data collected from mobile (AWARE framework), wearable sensors (Fitbit Flex2 and Inspire 2) and survey data (Ecological Momentary Assessment). Tasks associated with this dataset include:
\begin{itemize}[itemsep=0.05em,leftmargin=*]
\item \textbf{Depression (DEP)}: Use of machine learning algorithms that analyze patterns in user behavior and language use for depression detection.
\item \textbf{Anxiety (ANX)}: Identification of anxiety often relies on behavioral markers such as irregular sleep patterns or heightened physiological responses like increased heart rate.
\end{itemize}

\paragraph{\textbf{AW\_FB} \cite{DVN/ZS2Z2J_2020}} A dataset examining the accuracy of consumer wearable devices (GENEActiv, Apple Watch Series 2 and Fitbit Charge HR2) collected from $n=49$ participants (26 females and 23 males) over 104 hours of activity logs in a lab based protocol. Tasks associated with this dataset include:
\begin{itemize}[itemsep=0.05em,leftmargin=*]
\item \textbf{Calorie Burn (CAL)}: Estimation of the amount of energy expended by an individual during physical activities.
\item \textbf{Activity (ACT)}: Identification of the types of physical activities based on sensor data.
\end{itemize}

\begin{table*}[!t]
\centering
\tiny
\caption{\textbf{Performance Evaluation of LLMs on Health Prediction Tasks.} \textbf{STRS}: Stress, \textbf{READ}: Readiness, \textbf{FATG}: Fatigue, \textbf{SQ}: Sleep Quality, \textbf{SR}: Stress Resilience, \textbf{SD}: Sleep Disorder, \textbf{ANX}: Anxiety, \textbf{DEP}: Depression, \textbf{ACT}: Activity, \textbf{CAL}: Calories. ``-'' denotes the failure cases due to token size limit or unreasonable responses. ``N/A'' denotes the case where the prediction is not reported or cannot be conducted. For each column (task), the best result is \textcolor{darkgreen}{\textbf{bolded}}, and the second best is \underline{underlined}. CoT denotes the chain-of-thoughts and SC denotes the self-consistency prompting. For each task, arrows in the parenthesis indicate the desired direction of improvement. $\uparrow$ indicates higher values are better for accuracy, while $\downarrow$ indicates lower values are better for mean absolute error.}
\setlength{\tabcolsep}{4.0pt}
\begin{tabular}{p{0.1cm} p{2.37cm} p{1.085cm} p{1.085cm} p{1.085cm} p{1.085cm} p{1.085cm} p{1.085cm} p{1.085cm} p{1.085cm} p{1.085cm} p{1.085cm} p{1.085cm} p{1.085cm} p{1.085cm}}
\toprule
\multicolumn{1}{c}{} & \multicolumn{1}{c}{\textbf{Dataset}} & \multicolumn{4}{c}{\textbf{PMData}} & \multicolumn{2}{c}{\textbf{LifeSnaps}} & \multicolumn{2}{c}{\textbf{GLOBEM}} & \multicolumn{2}{c}{\textbf{AW\_FB}} \\
\cmidrule(lr){3-6} \cmidrule(lr){7-8} \cmidrule(lr){9-10} \cmidrule(lr){11-12} 
 \textbf{} & \textbf{Model} & \textbf{\makecell{STRS \\ ($\downarrow$)}} & \textbf{\makecell{READ \\ ($\downarrow$)}} & \textbf{\makecell{FATG \\ ($\uparrow$)}} & \textbf{\makecell{SQ \\ ($\downarrow$)}} & \textbf{\makecell{SR \\ ($\downarrow$)}} & \textbf{\makecell{SD \\ ($\uparrow$)}} & \textbf{\makecell{ANX \\ ($\downarrow$)}} & \textbf{\makecell{DEP \\ ($\downarrow$)}} & \textbf{\makecell{ACT \\ ($\uparrow$)}} & \textbf{\makecell{CAL \\ ($\downarrow$)}} \\
\midrule
\multirow{12}{*}{\rotatebox{90}{Zero-shot}} & MedAlpaca & 0.76 \fontsize{4.5pt}{4.5pt}\selectfont{\textpm \ 0.1} & 2.18 \fontsize{4.5pt}{4.5pt}\selectfont{\textpm \ 0.1} & 46.8 \fontsize{4.5pt}{4.5pt}\selectfont{\textpm \ 11} & 0.68 \fontsize{4.5pt}{4.5pt}\selectfont{\textpm \ 0.0} & 1.17 \fontsize{4.5pt}{4.5pt}\selectfont{\textpm \ 0.1} & 40.3 \fontsize{4.5pt}{4.5pt}\selectfont{\textpm \ 1.6} & 1.23 \fontsize{4.5pt}{4.5pt}\selectfont{\textpm \ 0.3} & 0.89 \fontsize{4.5pt}{4.5pt}\selectfont{\textpm \ 0.1} & 21.7 \fontsize{4.5pt}{4.5pt}\selectfont{\textpm \ 4.4} & 35.0 \fontsize{4.5pt}{4.5pt}\selectfont{\textpm \ 6.0}   \\
\end{tabular}
\end{table*}

\subsection{Models}

We experimented with multiple LLMs with different sizes, pre-training targets, and availability.

\begin{itemize}[itemsep=0.05em,leftmargin=*]
  \item \textbf{MedAlpaca} (7B, 13B) \cite{han2023medalpaca}: An advanced LLM, fine-tuned specifically for medical question-answering. Built upon the foundations of Alpaca, it utilizes a diverse array of medical texts.
  \item \textbf{PMC-Llama} (13B) \cite{wu2023pmcllama}: A specialized open-source LLM, incorporating knowledge from 4.8M biomedical papers and 30K medical textbooks. 
  \item \textbf{Asclepius} (7B) \cite{kweon2023publicly}: A publicly shareable clinical LLM developed to circumvent privacy regulations by training on synthetic clinical notes extracted from biomedical literature.  
  \item \textbf{ClinicalCamel} (70B) \cite{toma2023clinical}: An open LLM, fine-tuned on the LLaMA-2 70B architecture using QLoRA.
  \item \textbf{Flan-T5} (3B) \cite{chung2022scaling}: An instruction fine-tuned version of T5 or Text-to-Text Transfer Transformer Language Model. 
  \item \textbf{Palmyra-Med} (20B) \cite{unknown}: An LLM fine-tuned on a custom medical dataset, demonstrating superior performance on medical knowledge datasets like PubMedQA and MedQA. 
  \item \textbf{BioMedGPT} (7B) \cite{zhang2024biomedgpt}: An open multi-modal generative pre-trained transformer (GPT) for biomedicine, which closes the gap between natural language modality and diverse biomedical data modalities via large generative language models.
  \item \textbf{BioMistral} (7B) \cite{labrak2024biomistral}: A Mistral-based further pre-trained open source model (\texttt{BioMistral 7B DARE}) suited for the medical domains.
  \item \textbf{Llama 2} (7B) \cite{touvron2023llama}: Meta AI's open LLM, which possess features of including a longer context length, improved training on a larger dataset, and fine-tuning via RLHF for greater performance.
  \item \textbf{GPT-3.5} (175B) \cite{gpt-3.5}: A variant of GPT-3.5 (\texttt{gpt-3.5-} \texttt{turbo-instruct}) from OpenAI, specifically fine-tuned to provide direct answers or complete text rather than simulating conversations. 
  \item \textbf{GPT-4} \cite{openai2023gpt4}: A state-of-the-art autoregressive language model from OpenAI (\texttt{gpt-4}) exhibiting remarkable capabilities in various NLP tasks including translation, question answering, and text generation without task-specific fine-tuning.
  \item \textbf{Gemini-Pro} \cite{geminiteam2023gemini}: Google DeepMind's versatile model optimized for diverse tasks across text, code, images, and audio. It offers a balance between capability and efficiency, suitable for both high-performance applications and on-device tasks, making advanced AI accessible to developers and enterprises alike. 
  \item \textbf{Baseline}: We also compare LLMs against a set of baseline techniques, including simple majority class (classification only), traditional ML models (MLP, SVM, RandomForest), and traditional pre-trained language models (BERT, BioMed-RoBERTa, and PALM$_{\text{p-tuned}}$). Note that these baseline models are all task-specific and thus need to be trained on data from target tasks.
\end{itemize}

\begin{table*}[t!]
  \centering
  \footnotesize
  \caption{\textbf{Balanced Cross-Dataset Performance Summary of MedAlpaca-7b Finetuning on Single Dataset.} Numbers indicate the results of the model finetuned and tested on the same dataset. The bottom four rows are related Alpaca versions for reference. \textcolor{GREEN}{Green}/\textcolor{RED}{Red} color marks the ones with better/worse cross-dataset performance compared to the zero-shot version MedAlpaca. \boxit{0.24in}{.076in}Box \ indicate the results of the case where the model fine-tuned and tested on the same dataset. For each task, arrow in the parenthesis indicate the desired direction of improvement. $\uparrow$ indicates higher values are better for accuracy, while $\downarrow$ indicates lower values are better for mean absolute error.}
  \label{tab:mytable}
  \begin{tabular}{lcccccccccc}
    \toprule
    \textbf{Test Dataset} & \multicolumn{4}{c}{\textbf{PMData}} & \multicolumn{2}{c}{\textbf{LifeSnaps}} & \multicolumn{2}{c}{\textbf{GLOBEM}} & \multicolumn{2}{c}{\textbf{AW\_FB}} \\
    \cmidrule(lr){2-5} \cmidrule(lr){6-7} \cmidrule(lr){8-9} \cmidrule(lr){10-11}
    \textbf{Finetune Dataset} & \makecell{\textbf{STRS} \\ ($\downarrow$)} & \makecell{\textbf{READ} \\ ($\downarrow$)} & \makecell{\textbf{FATG} \\ ($\uparrow$)} & \makecell{\textbf{SQ} \\ ($\downarrow$)} & \makecell{\textbf{SR} \\ ($\downarrow$)} & \makecell{\textbf{SD} \\ ($\uparrow$)} & \makecell{\textbf{ANX} \\ ($\downarrow$)} & \makecell{\textbf{DEP} \\ ($\downarrow$)} & \makecell{\textbf{ACT} \\ ($\uparrow$)} & \makecell{\textbf{CAL} \\ ($\downarrow$)} \\
    \midrule
    PMData & \boxit{1.78in}{.06in}\underline{0.38} &  \textbf{0.94}  &  \textbf{71.4} & 0.90 &  \textcolor{RED}{2.02} & \textcolor{GREEN}{72.1} & \textcolor{RED}{2.01} & \textcolor{RED}{1.76} & - \\
    LifeSnaps & \textcolor{GREEN}{0.71} & \textcolor{RED}{3.26} & - & - & \boxit{0.62in}{.060in}\textbf{0.45} &  80.0 & \textcolor{GREEN}{0.86} & \textcolor{RED}{0.62} & \textcolor{GREEN}{25.0} & \textcolor{GREEN}{34.8}  \\
    GLOBEM & \textcolor{GREEN}{0.72} & \textcolor{RED}{4.62} & \textcolor{RED}{23.8} & \textcolor{RED}{1.25} & \textcolor{GREEN}{0.72} & \textcolor{GREEN}{\underline{81.2}} & \boxit{0.69in}{.06in}\underline{0.81} & \textbf{0.29} & - & \textcolor{GREEN}{33.9}  \\
    AW\_FB & \textcolor{GREEN}{0.40} & \textcolor{GREEN}{1.94} & - & \textcolor{RED}{1.37} & \textcolor{GREEN}{0.94} & \textcolor{GREEN}{\textbf{93.3}} & \textcolor{RED}{1.91} & \textcolor{RED}{1.57} & \boxit{0.68in}{.06in}\textbf{43.5} & 33.2  \\ 
    \midrule
    \textbf{Reference}  \\
    \midrule
    MedAlpaca\_{ZS} & 0.76 & 2.18 & 46.8 & 0.68 & 1.17 & 40.3 & 1.23 & 0.89 & 21.7 & 35.0  \\
    MedAlpaca\_{FS}$^{*}$ & 0.50 & 1.94 & 36.2 & 0.51 & 0.94 & 49.6 & 0.97 & 0.56 & 19.3 & 35.3 \\
    MedAlpaca\_{T} & 0.53 & 1.38 & 58.1 & \underline{0.50} & 0.68 & 73.3 & 0.98 & 0.81 & 16.1 & \textbf{6.52} \\
    HealthAlpaca & \textbf{0.31} & \underline{1.32} & \underline{70.7} & \textbf{0.35} & \underline{0.62} & 72.1 & \textbf{0.46} & \underline{0.49} & \underline{41.7} & \underline{31.5} \\
    \bottomrule
  \end{tabular}
  
  \footnotesize{ZS: Zero Shot Prompting} \\
  \footnotesize{FS$^*$: Best Performance among Few Shot Prompting} \\
  \footnotesize{T: Task-specific Fine-tuning} \\
  \label{tab:dataset_specific_finetuning}
\end{table*}

\vspace{-4px}

\section{Results and Discussion}
\label{sec:floats}

Table \ref{tab:main_table} (w/ MAE and Accuracy metric) and Table \ref{tab:main_table2} (w/ MAPE and F1-score metric) in Appendix shows the performance of twelve off-the-shelf LLMs prompted/trained with a set of prompting/fine-tuning techniques and six task-specific baseline models across ten health prediction tasks and four health datasets. 
Also, Tables \ref{app:ttest1}-\ref{app:ttest10} in Appendix illustrates the paired t-test results conducted between our fine-tuned models and the baselines.

\subsection{Zero-shot and Few-shot Performance}



In the \textit{zero-shot} setting, there was no clear superiority among the models and surprisingly, relatively smaller models (number of parameters less than 100B) showed better performance (GPT series models and Gemini-Pro took proportion of only 20\% among best and second best performers) than GPT series models and Gemini-Pro. This may suggest that the effectiveness of a model in the health domain is not solely dependent on its size but could also be significantly influenced by its pre-training knowledge in a zero-shot setting. Asclepius for example, was pre-trained on 158k high-quality synthetic clinical dataset, performed on par with GPT-3.5-turbo showing task-specific knowledge can potentially outweigh the benefits of sheer model size \cite{kweon2023publicly}.\\In the few-shot setting however, GPT-3.5 and GPT-4 showed improvement with \textit{few-shot} prompting along with CoT and SC in most of the tasks, while such improvement is not significantly observed with smaller models. Specifically, CoT-SC greatly enhanced model's performance by showing the best performance with CoT-SC in 9 out of 10 tasks indicating the benefit of structured prompting in understanding and predicting behavior. This finding is aligned with prior findings from other non-health tasks in \cite{tay2022unifying, suzgun2022challenging}. This indicates that large LLMs have a stronger capability of quickly learning from examples for health tasks.

\vspace{-4px}

\subsection{Finetuning Performance}

 Across all categories, our fully \textit{fine-tuned} model, HealthAlpaca, shows the best performance in 8 out of 10 consumer health tasks (Table \ref{tab:main_table2}). HealthAlpaca achieves comparable or better performance than GPT-series models and Gemini-Pro which has a two-magnitude larger number of parameters. HealthAlpaca-lora, a model \textit{fine-tuned} with a parameter-efficient finetuning technique LoRA, also shows a performance boost over the larger models in almost all tasks. These results suggest that LLMs can be effectively tuned for tasks with multi-modal time-series wearable data.\\ Also, as predicted, HealthAlpaca-13b showed better performance than HealthAlpaca-7b and parameter-efficiently fine-tuned models in most of the tasks as it involved more number of parameters and updates in those larger number of parameters.
\label{subsec:ablation1}

\subsection{Generalization of Fine-tuned Models across Datasets}
To assess the generalization capabilities of our \textit{fine-tuned} models, we conducted dataset-specific fine-tuning. We then compared model performance with reference results from task-specific and multi-dataset \textit{fine-tuned} models.
The results presented in Table \ref{tab:dataset_specific_finetuning} indicate that not surprisingly, MedAlpaca, when fine-tuned on the same dataset with the target task mostly shows the best performance compared to when fine-tuned on other datasets.
We also observed that while dataset-specific fine-tuning often failed to predict tasks from other datasets, the multi-dataset \textit{fine-tuned} HealthAlpaca, exhibited reasonable generalization performance across tasks.

Moreover, we made an intriguing observation regarding the synergistic effect between certain datasets. In a few cases, such as AW\_FB $\rightarrow$ STRS, and LifeSnaps $\rightarrow$ ANX, the performance of the \textit{fine-tuned} models surpassed that of the zero-shot and dataset-specific fine-tuning approaches. These findings suggest that fine-tuning on a single dataset can provide health knowledge to a certain extent and thereby improve overall generalization results. However, such improvement is not consistently observed across all tasks, and it depends on the overlapping content across datasets.

\begin{figure*}[t!]
    \vspace{-0.5cm}
    \centering
    \includegraphics[width=0.9\linewidth]{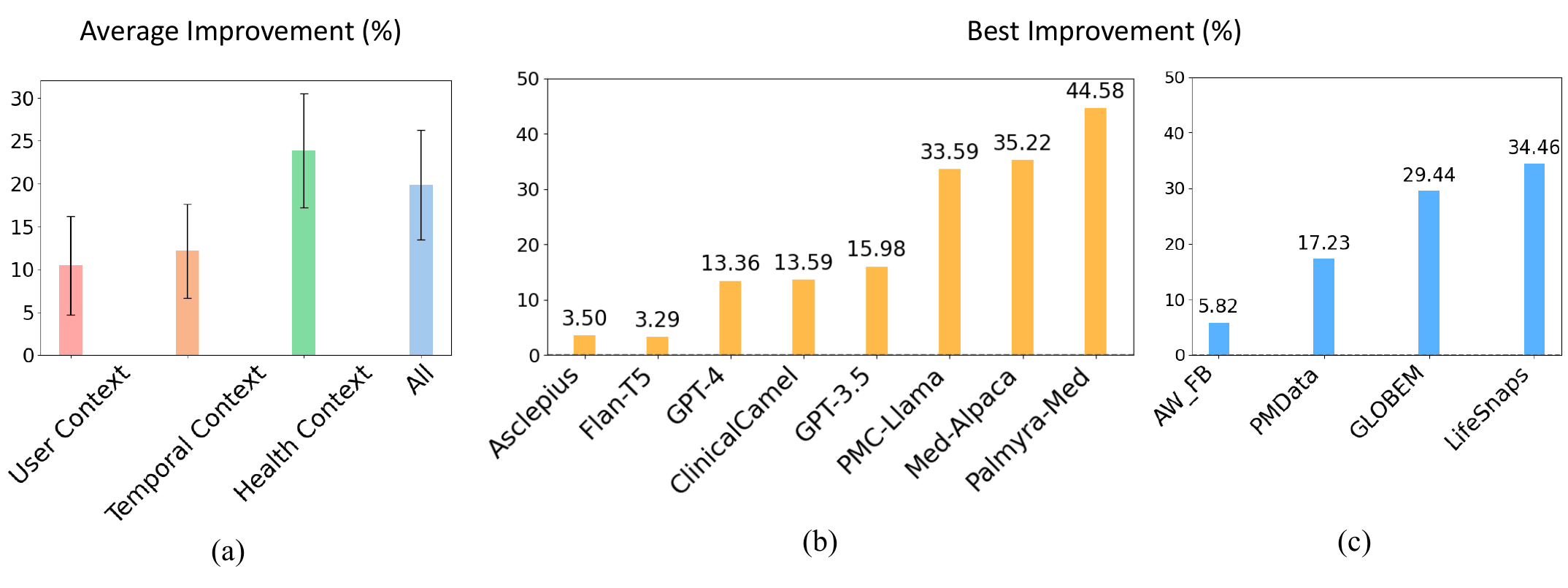}
    \caption{(\textbf{a}): Average Performance Improvement over basic (\textit{bs}) across contexts. (\textbf{b}): Best Performance Improvement across LLMs. (\textbf{c}): Best Performance Improvement across Datasets. Note that few models (Llama 2, Gemini-Pro, BioMedGPT and BioMistral) were excluded in this experiment due to the prioritization of models based on integration timelines.}
    \label{fig:balanced_models}
\end{figure*}

\subsection{Importance of different contexts in Prompt Designing for Healthcare LLMs}
In Figure \ref{fig:balanced_models}, we present the summarized effect of context enhancement strategies in \textit{zero-shot} setting across LLMs and datasets. Overall, adding contexts can significantly improve the model performance. Among four types of context information, adding health context shows the biggest performance boost in average. More interestingly, our results show varied effectiveness on different LLMs and datasets.
From an LLM perspective (shown in Figure \ref{fig:balanced_models}b), Palmyra-Med benefited the most from the enhancement, where it shows up to 44.58\% improvement when adding temporal context ($tc$). Palmyra-Med's significant improvement can be attributed to its fine-tuning with two medical datasets: PubMedQA, a resource focused on question-answering in the medical domain, and MedQA, which provides a broad range of medical questions and answers. By integrating temporal context, Palmyra-Med leverages its enhanced medical literature comprehension and temporal pattern recognition, leading to substantial improvements in predictive accuracy by effectively interpreting medical conditions' progression and patterns over time. However, $tc$ does not help improve the performance for large-size LLMs like GPT-3.5 or GPT-4. This is probably because they already possess the capability to understand the statistical traits in the time-series data as shown in the case study in Figure \ref{fig:case_study_readiness} (e.g. average, max, variation).
From the dataset perspective, our enhancement strategies boost performance in all datasets, and LifeSnaps shows the best improvement up to 34.46\%. 
GLOBEM and AW\_FB showed significant performance improvement when applying $tc$ and $all$ context enhancements. This might be because of temporal aspects in the dataset. For instance, GLOBEM provides 7- and 14-days statistics for each feature, which highlights the temporal dimension. 

\label{subsec:ablation1}

\subsection{Importance of Training Size in Fine-tuning Performance}
To understand the required amount of data for effective fine-tuning, we conduct experiments on fine-tuning Alpaca with varying down-sampled training sizes: 5\%, 15\%, 25\%, 50\%, and 100\% of the original dataset, and evaluate health prediction performances with three seeds. Figure \ref{fig:train_size} in Appendix shows the overall results. With 15\% of the original dataset, the fine-tuned model already outperforms the \textit{zero-shot} performance on all 13 tasks.
This number provides guidance for fine-tuning when computing resources are limited.
As expected, we observe an increasing trend in performance with more training data. 

\vspace{-4pt}

\subsection{Case Study of LLM's Capability on Health Reasoning}

In addition to health predictions, it is critical to see whether the health reasoning are also valid especially in healthcare domain. Here, we show two representative cases; 1) READ: Readiness Score Prediction and 2) SD: Sleep Disorder Prediction across four LLMs including our fine-tuned model HealthAlpaca.  

\vspace{-4pt}

\subsubsection{LLM's capability to understand the time-series data}

In analyzing time-series data, LLMs show distinct approaches. HealthAlpaca for example, shows a conservative interpretation of data like steps and mood scores to suggest moderate readiness levels (see Figure \ref{fig:case_study_readiness}). It emphasizes stability, suggesting suitability for moderate activities. GPT-4 on the other hand, identifies days with high activity to suggest greater readiness, highlighting its focus on peak performance. In sleep disorder assessments, HealthAlpaca indicates no sleep disorder, looking at overall sleep metrics for a broad assessment. GPT-3.5 uses a formulaic approach focusing on averages, which might overlook deeper sleep quality nuances. GPT-4 points out inconsistencies and lower sleep efficiency, suggesting potential sleep issues, demonstrating its attention to variability. This comparison reveals how different models apply inference of health data, from HealthAlpaca's broad, stable approach to GPT-4's detailed, variable-sensitive analysis.

\subsubsection{False and Hallucinated Reasoning from LLMs}

False reasoning in LLMs occurs when data is misinterpreted or generic standards are misapplied. In readiness evaluations (Figure \ref{fig:case_study_readiness}), such reasoning can lead to overestimation or underestimation of physical readiness. For instance, HealthAlpaca might not fully value high-activity days, while GPT-4 could underestimate the need for consistent activity. In case of GPT-3.5, there was a misinterpretation of the question, particularly around the provided resting heart rate information. Gemini-Pro, on the other hand, makes reasonable predictions but misinterprets that the average calorie burns are in the range of recommended daily caloric burn. In sleep disorder assessments (Figure \ref{fig:case_study_sleep_disorder}), divergent model conclusions highlight the challenge of accurate health reasoning. HealthAlpaca might miss nuanced sleep patterns, while GPT-3.5's straightforward analysis could overlook sleep quality's complexity. This emphasizes the need for nuanced interpretation and the integration of detailed variability analysis, as seen with GPT-4. These examples show the critical balance LLMs must achieve between detailed data analysis and overarching health perspectives to provide reliable health assessments.

\vspace{-4pt}

\section{Conclusion}
\label{sec:cite}

In this paper, we present the first comprehensive evaluation of twelve off-the-shelf LLMs across ten consumer health prediction tasks (binary, multi-class classification, and regression) spanning four public health datasets. Our experiments encompass a variety of prompting and fine-tuning techniques. The results reveal several interesting findings. First, our context enhancement strategy boosts the performance across all datasets and LLMs, particularly emphasizing the importance of incorporating health knowledge context in prompts. More importantly, our fine-tuned model, HealthAlpaca, demonstrates the best performance in 8 out of 10 tasks, outperforming much larger models such as GPT-4 and Gemini-Pro, even when these are equipped with few-shot prompting. Additionally, we conducted a case study on selected examples to highlight the LLMs' reasoning capabilities and limitations regarding false and hallucinated reasoning in health predictions. Ethical concerns regarding privacy and bias still remains, necessitating further investigation before real-world deployment.

\paragraph{Limitation and Future Work} This study's reliance on self-reported health data limits its clinical applicability \cite{zack2023coding} and raises ethical considerations, particularly regarding data validity and user communication. Additionally, the ``black-box" nature of LLMs complicates the assessment of their clinical validity. To address these issues, future work will focus on: 1) conducting evaluations with clinically diagnosed datasets in collaboration with healthcare professionals to enhance clinical relevance; 2) ensuring ethical and regulatory compliance, particularly in how health-related predictions are communicated to users; 3) improving LLMs' explainability to facilitate understanding of their decision-making processes, thereby aiding in the accurate interpretation of health predictions; and 4) incorporating privacy-preserving technologies like federated learning \cite{raeini2023privacy} to protect sensitive health information.



\acks{We thank Yoon Kim and Rosalind Picard at MIT, Vivek Natarajan and Ming-Zher Poh at Google for their revisions, feedback, and support.}

\bibliography{main}

\tikzstyle{mybox} = [draw=black, very thick,
    rectangle, rounded corners, inner sep=10pt, inner ysep=13pt]
\tikzstyle{fancytitle} =[fill=black, text=white]

\appendix

\newpage

\section{Implementation Details}\label{apd:first}
We fine-tune our Health-LLMs on 4 NVIDIA A100 80GB GPUs with a batch size of 128 with different number of epochs for the purpose of fine-tuning. We based on MedAlpaca-7b and -13b models and conducted instruction fine-tuning with cross entropy as the loss function, we backpropagate and update model parameters in 3 and 5 epochs (7b and 13b respectively), with Adam optimizer and a learning rate as $2e^{-5}$ (consine learning rate scheduler and warmup steps of 100). It took about 3.4 hours for 5 epochs of training with the default training setting. For more training details, we follow the default parameters in the original code repository of MedAlpaca\footnote{https://github.com/kbressem/medAlpaca}. We set default decoding method as sampling and also use other decoding methods such as controlling the temperature in sampling to generate different reasoning paths. The codes and fine-tuned models will be made publicly available upon the release of the camera-ready version of this paper. \\For the zero-shot and few-shot prompting, we utilized open-source models from huggingface with four Nvidia A6000 GPUs and used OpenAI and GenAI APIs for the closed-source models. We conducted oversampling for each dataset to resolve class imbalance issue and merged the train sets together. For the task-specific and multi-dataset used for fine-tuning, we will upload the codes to generate these datasets in the code repository by camera-ready version.

\section{Temporal Encoding Methods} \label{apd:second}

\subsubsection{\textbf{Natural Language String}} Following the approach presented in \cite{gruver2023large}, we also transform raw time-series data into a language string. This method, known for its simplicity and effectiveness, converts sequential data into an understandable format for both humans and AI models. For instance, a 14-day sensor reading for a consumer health prediction task can be transformed into daily readings of steps, and resting heart rate (\{Nan, 991.0, ..., Nan\}). 

\subsubsection{\textbf{Modality-specific Encoding}} In line with the methodology proposed in \cite{belyaeva2023multimodal}, this method utilizes pre-trained encoders for each modality to embed non-text data modalities including time-series data into the same latent space as the language tokens. Though this method may provide nuanced data representations, it introduces complexity and computational overhead that may not always yield performance gains.

\subsubsection{\textbf{Statistical Summary}} Similar to the methods used in \cite{jin2023timellm}, this method encapsulates time-series data into statistical summaries such as mean, std, median, etc. While this approach reduces the volume of data to be processed, it may risk overlooking critical temporal patterns inherent in the raw data.\\

\section{Additional Experiments}

\begin{table}[ht]
\centering
\footnotesize
\begin{tabular}{lcc}
\toprule
\textbf{\makecell{Fine-tuned\\Dataset}} & \multicolumn{2}{c}{\textbf{Tested on LifeSnaps}} \\
\cmidrule{2-3}
& \textbf{STRS} & \textbf{CAL} \\
& (MAE ± SD) & (MAE ± SD) \\
\midrule
PMData & 0.44 ± 0.1 & - \\
AW\_FB & - & 44.2 ± 7.2 \\
\bottomrule
\end{tabular}
\caption{Summary of additional experiments showing the performance of HealthAlpaca-7b on the LifeSnaps dataset for stress level (STRS) and calorie burn (CAL) prediction.}
\label{tab:additional_experiments}
\end{table}

Additional experiments were conducted to address concerns regarding potential dataset overlap between fine-tuning and evaluation datasets, potentially favoring HealthAlpaca unfairly. 
Specifically, we focused on 1) stress level (STRS) prediction and 2) calorie burn (CAL) prediction using LifeSnaps dataset labels, which were fine-tuned on PMData and AW\_FB datasets. These experiments were designed to showcase HealthAlpaca's generalizability and robustness by entirely separating the training and evaluation datasets. Results demonstrated promising performance with a mean absolute error (MAE) of 0.44 ± 0.1 for stress level prediction and MAE of 44.2 ± 7.2 for calorie burn prediction. Furthermore, we plan to extend these findings by incorporating additional experimental results with HealthAlpaca-7b-lora, HealthAlpaca-13b-lora, and HealthAlpaca-13b models. Additionally, future work will include more cross-dataset experiments using the GLOBEM dataset to further evaluate HealthAlpaca's performance across diverse datasets.


\begin{figure*}[b] 
    \centering
    \includegraphics[width=0.85\linewidth]{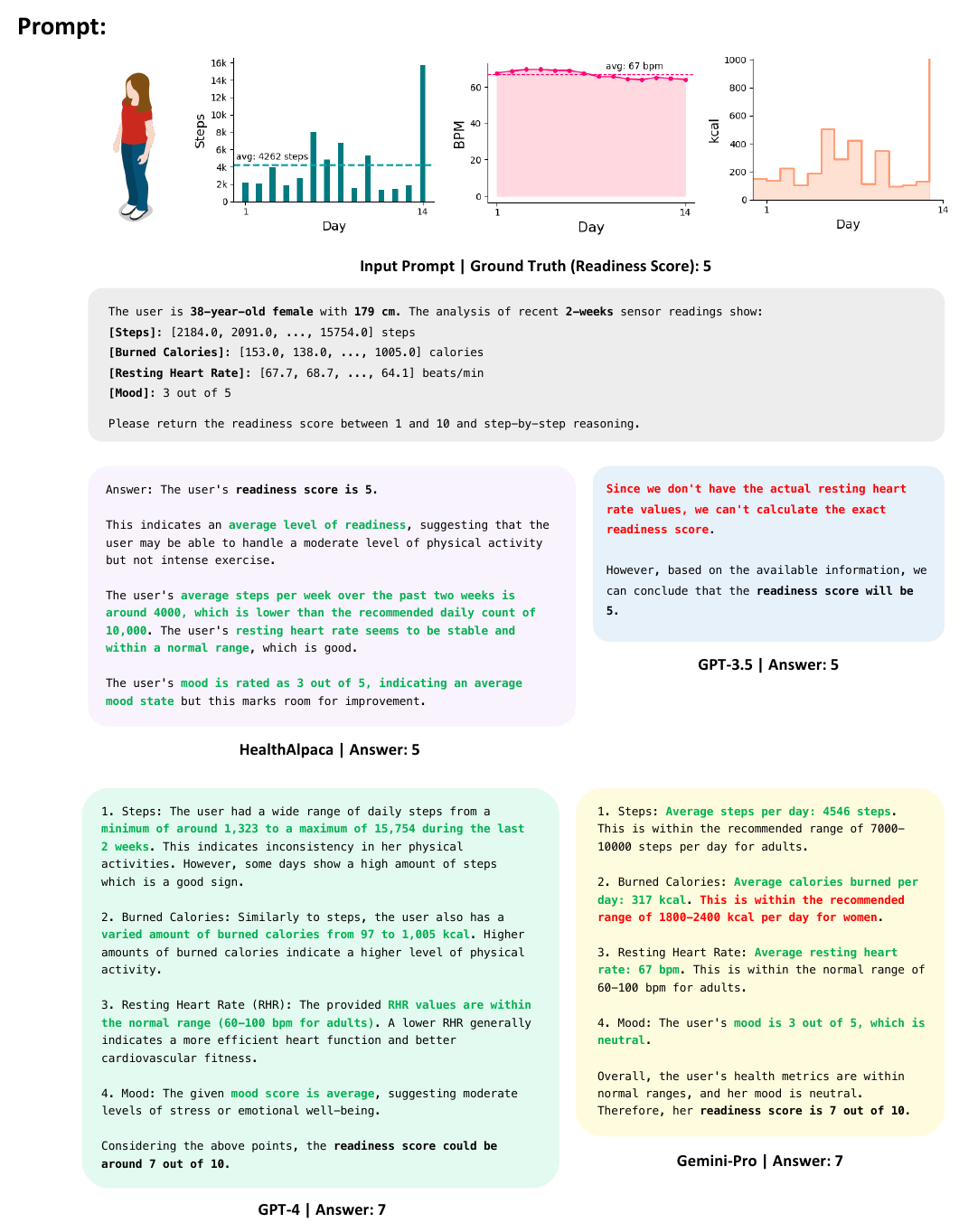}
    \caption{A Case Study on Readiness Score Prediction (READ) from PMData dataset. Here, we display the responses from 1) our fine-tuned model, HealthAlpaca, 2) GPT-3.5, 3) GPT-4 and 4) Gemini-Pro. \textcolor{darkgreen}{\textbf{Green Bolded texts}} highlights the valid reasoning and \textbf{Red Bolded texts} highlights the false or irrelevant reasoning to the input.}
    \label{fig:case_study_readiness}
\end{figure*}

\begin{figure*}[b] 
    \centering\includegraphics[width=0.85\linewidth]{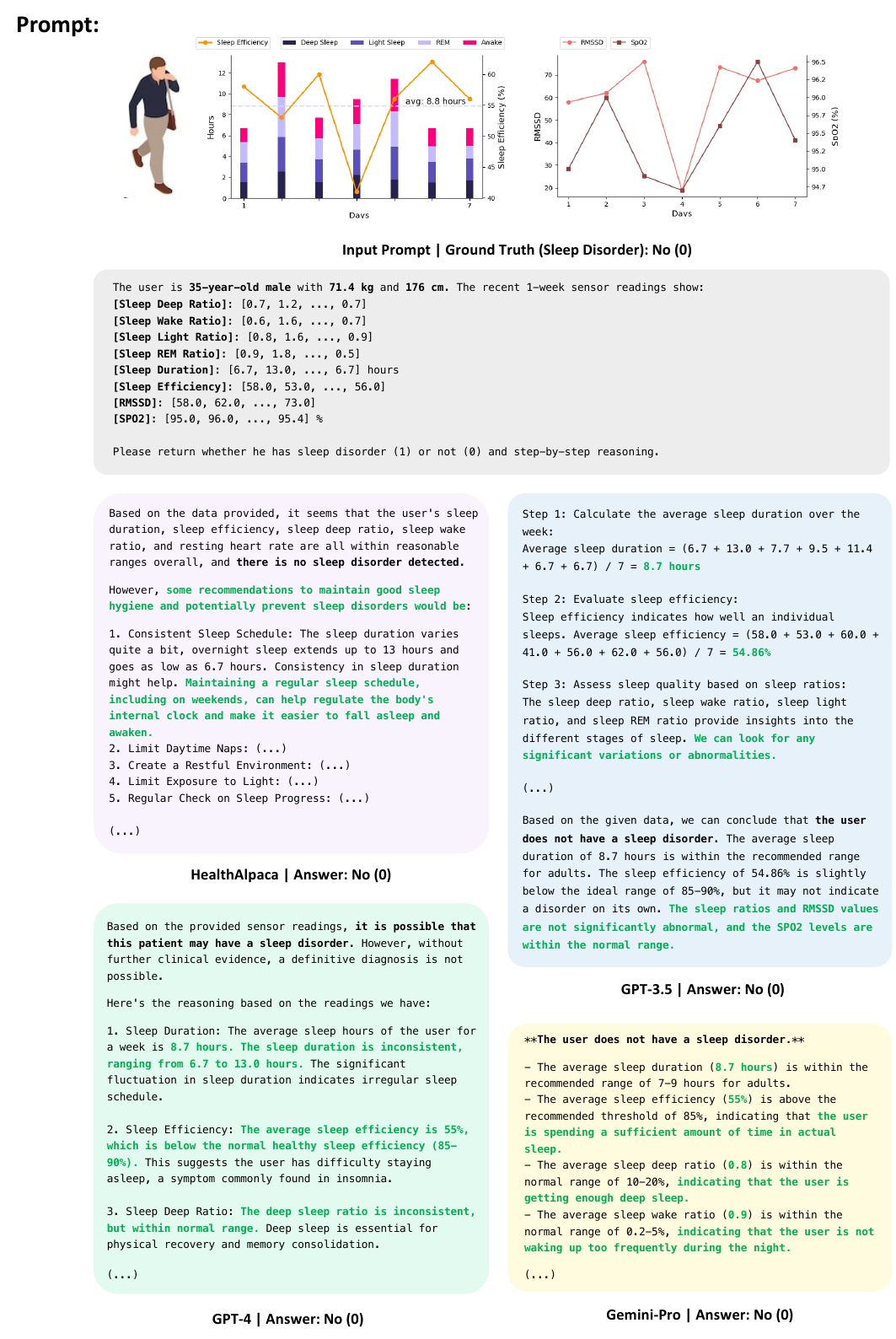}
    \caption{A Case Study on Sleep Disorder Prediction (SQ) from LifeSnaps dataset. Here, we display the responses from 1) our fine-tuned model, HealthAlpaca, 2) GPT-3.5, 3) GPT-4 and 4) Gemini-Pro. \textcolor{darkgreen}{\textbf{Green Bolded texts}} highlights the valid reasoning.}
    \label{fig:case_study_sleep_disorder}
\end{figure*}

\definecolor{RoyalBlue}{cmyk}{1, 0.90, 0, 0}

\begin{figure*}[t!]
    \centering
    \includegraphics[width=1.0\linewidth]{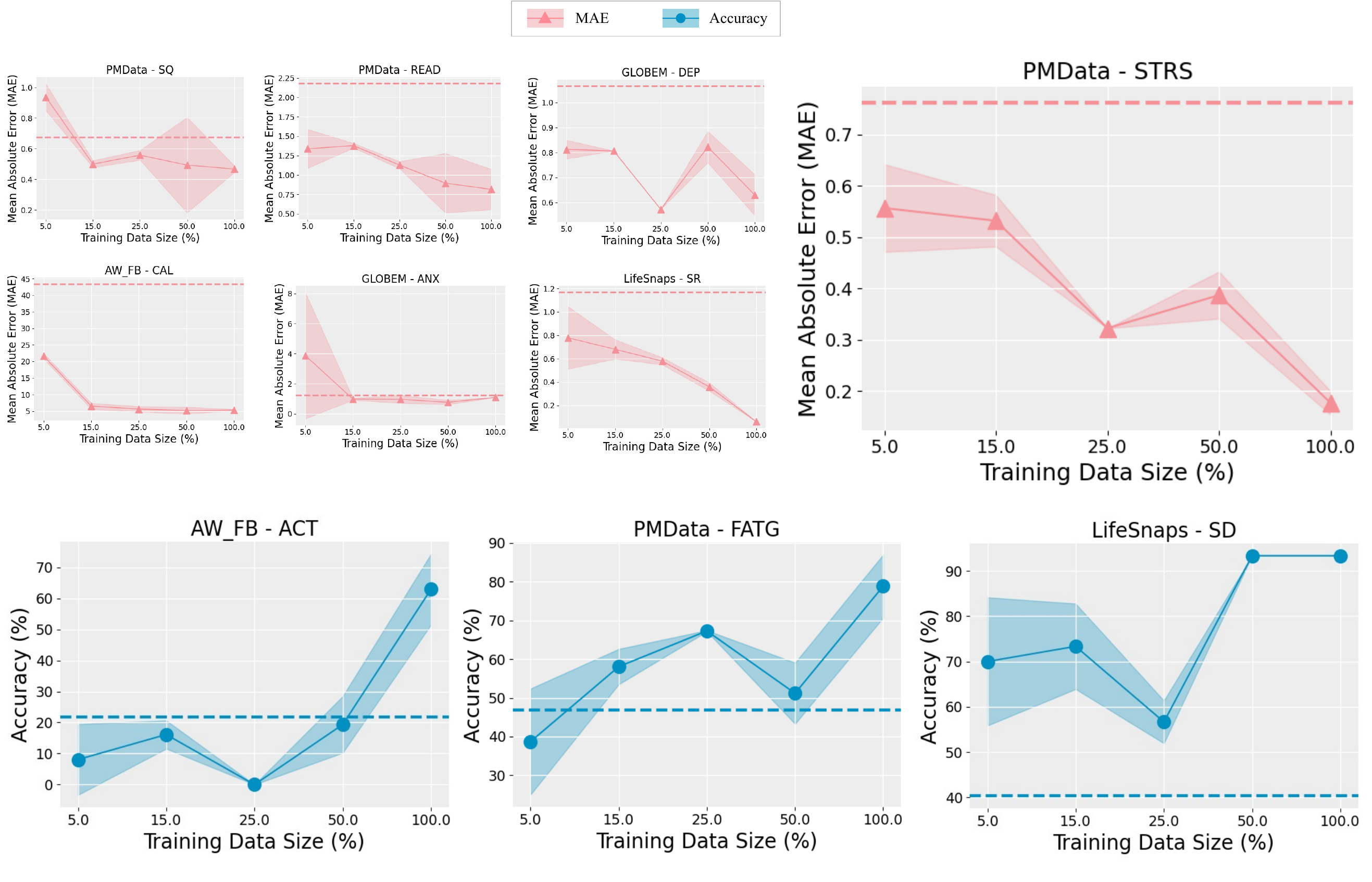}
    \caption{\textbf{Health Prediction Performance of Fully \textit{fine-tuned} MedAlpaca with Different Training Sizes.} The instruction fine-tuning is conducted across ten tasks across four datasets. The solid lines represents the \textit{fully fine-tuned} model's performance whereas the dashed lines represents the \textit{zero-shot} performance of MedAlpaca which serves as baselines. Note that the color indicates the metrics used to evaluate the tasks.} 
    \label{fig:train_size}
\end{figure*}

\tikzstyle{mybox} = [draw=black, very thick,
    rectangle, rounded corners, inner sep=10pt, inner ysep=13pt]
\tikzstyle{fancytitle} =[fill=black, text=white]

\sethlcolor{black}

\begin{figure*}[ht!]
\caption{Time-series Data Encoding Methods.}
\centering

\begin{tabular}{c} 
\begin{tikzpicture}
\node [mybox] (box){%
    \begin{minipage}{0.935\textwidth}
    \footnotesize
        \begin{flushleft} 
        The analysis of recent period: \{14\} days sensor readings show: 
        
        Steps: \tcbox[on line, boxsep=1pt, left=0pt,right=0pt,top=0pt,bottom=0pt,colframe=tc_color!30, colback=tc_color!30]{\{``NaN, 991.0, ..., NaN"\}} steps, Burned Calories: \tcbox[on line, boxsep=1pt, left=0pt,right=0pt,top=0pt,bottom=0pt,colframe=tc_color!30, colback=tc_color!30]{\{``NaN, 94.0 ..., NaN"\}} kcal, Resting Heart Rate: \tcbox[on line, boxsep=1pt, left=0pt,right=0pt,top=0pt,bottom=0pt,colframe=tc_color!30, colback=tc_color!30]{\{``69.32, 67.72, ..., 64.55"\}} beats/min, Sleep Minutes: \tcbox[on line, boxsep=1pt, left=0pt,right=0pt,top=0pt,bottom=0pt,colframe=tc_color!30, colback=tc_color!30]{\{``534.0, 455.0, ..., 405.0"\}} minutes, [Mood]: 3 out of 5. 

        \vspace{5pt}
        
        In this regard, what is the predicted readiness score/level between 0 and 10?
        \end{flushleft}
    \end{minipage}
};
\node[fancytitle, rounded corners, right=10pt] at (box.north west) {1. Natural Language String};
\label{fig:time-series_encoding1}
\end{tikzpicture}%
\end{tabular}

\vspace{10pt}

\begin{tabular}{c} 
\begin{tikzpicture}
\node [mybox] (box){%
    \begin{minipage}{0.935\textwidth}
    \footnotesize
        \begin{flushleft} 

        sentence = ``The analysis of recent period: \{14\} days sensor readings show: Steps: \tcbox[on line, boxsep=1pt, left=0pt,right=0pt,top=0pt,bottom=0pt,colframe=tc_color!30, colback=tc_color!30]{ $<\text{modality}_1>$ } steps, Burned Calories: \tcbox[on line, boxsep=1pt, left=0pt,right=0pt,top=0pt,bottom=0pt,colframe=tc_color!30, colback=tc_color!30]{$<\text{modality}_2>$} kcal, Resting Heart Rate: \tcbox[on line, boxsep=1pt, left=0pt,right=0pt,top=0pt,bottom=0pt,colframe=tc_color!30, colback=tc_color!30]{$<\text{modality}_3>$} beats/min, Sleep Minutes: \tcbox[on line, boxsep=1pt, left=0pt,right=0pt,top=0pt,bottom=0pt,colframe=tc_color!30, colback=tc_color!30]{$<\text{modality}_4>$} minutes, [Mood]: 3 out of 5. In this regard, what is the predicted readiness score/level between 0 and 10?"

        \vspace{5pt}

        \text{for $word$ in $sentence$:} 
        
        \begin{align*}
            &\quad w_i = \text{tokenizer}(word) \\
            &\quad x_i = \begin{cases}
                \gamma(w_i), & \text{if } i \text{ is a text token},\\
                \phi_m(w_i), & \text{if } i \text{ corresponds to the modality } m
            \end{cases}
        \end{align*}

        \vspace{5pt}

        where $\gamma : \mathcal{W} \rightarrow \mathcal{X}$ is the large embeding matrix and $\phi_m$: $\mathcal{M} \rightarrow \mathcal{X}^{q_m}$ is the pre-trained encoder for modality $m$. 
        
        \end{flushleft}
    \end{minipage}
};
\node[fancytitle, rounded corners, right=10pt] at (box.north west) {2. Modality-specific Encoding};
\label{fig:encoding_methods}
\end{tikzpicture}%
\end{tabular}

\vspace{10pt}

\begin{tabular}{c} 
\begin{tikzpicture}
\node [mybox] (box){%
    \begin{minipage}{0.935\textwidth}
    \footnotesize
        \begin{flushleft} 
        The analysis of recent 14 days of sensor readings show:
        
        Steps: \tcbox[on line, boxsep=1pt, left=0pt,right=0pt,top=0pt,bottom=0pt,colframe=tc_color!30, colback=tc_color!30]{\{``avg": 812, ``std": 102, ``min: 304", ``max": 1402\}} steps, 
        
        Burned Calories: \tcbox[on line, boxsep=1pt, left=0pt,right=0pt,top=0pt,bottom=0pt,colframe=tc_color!30, colback=tc_color!30]{\{``avg": 97, ``std": 43, ``min: 23", ``max": 100\}} kcal, 
        
        Resting Heart Rate: \tcbox[on line, boxsep=1pt, left=0pt,right=0pt,top=0pt,bottom=0pt,colframe=tc_color!30, colback=tc_color!30]{\{``avg": 66.54, ``std": 22.14 , ``min: 55", ``max": 85\}} beats/min, 
        
        Sleep Minutes: \tcbox[on line, boxsep=1pt, left=0pt,right=0pt,top=0pt,bottom=0pt,colframe=tc_color!30, colback=tc_color!30]{\{``avg": 487, ``std": 179, ``min: 320", ``max": 621\}} minutes, 
        
        Mood: 3 out of 5. 

        \vspace{5pt}
        
        In this regard, what is the predicted readiness score/level between 0 and 10?
        \end{flushleft}
    \end{minipage}
};
\node[fancytitle, rounded corners, right=10pt] at (box.north west) {3. Statistical Summary};
\end{tikzpicture}%
\end{tabular}

\end{figure*}

\begin{figure*}
\caption{Zero-shot, Few-shot and Few-shot CoT-SC exemplars for READ.}
\centering

\begin{tabular}{c} 
\begin{tikzpicture}
\node [mybox] (box){%
    \begin{minipage}{0.935\textwidth}
    \footnotesize
        \begin{flushleft} 
        {\textbf{Instruction:}} You are an intelligent healthcare agent. 
        
        {\textbf{Question:}} The user is 38-year-old female with 179 cm. The analysis of recent 2-weeks sensor readings show:
        
        \begin{itemize}
            \item Steps: [5260.0, 1323.0, ..., 3308.0] steps 
            \item Burned Calories: [349.0, 97.0, ..., 206.0] calories
            \item Resting Heart Rate: [65.61, 64.33, ..., 68.26] beats/min
            \item Sleep Minutes: [406.0, 533.0, ..., 462.0] minutes
            \item Mood: 3 out of 5
        \end{itemize}
        
        In this regard, can you predict user's readiness score between 0 and 5? \vspace{5pt}
        
        {\textbf{Response:}}
        \end{flushleft}
    \end{minipage}
};
\node[fancytitle, rounded corners, right=10pt] at (box.north west) {Prompt template for zero-shot learning};
\end{tikzpicture}%
\end{tabular}

\vspace{10pt}

\begin{tabular}{c} 
\begin{tikzpicture}
\node [mybox] (box){%
    \begin{minipage}{0.935\textwidth}
    \footnotesize
        \begin{flushleft} 
        {\textbf{ Instruction:}} You are an intelligent healthcare agent. \vspace{5pt}

        {\textbf{\{\{few\_shot\_examples\}\}}} \vspace{5pt}
        
        {\textbf{ Question:}} The user is 38-year-old female with 179 cm. The analysis of recent 2-weeks sensor readings show:
        
        \begin{itemize}
            \item Steps: [5260.0, 1323.0, ..., 3308.0] steps 
            \item Burned Calories: [349.0, 97.0, ..., 206.0] calories
            \item Resting Heart Rate: [65.61, 64.33, ..., 68.26] beats/min
            \item Sleep Minutes: [406.0, 533.0, ..., 462.0] minutes
            \item Mood: 3 out of 5
        \end{itemize}
        
        In this regard, can you predict user's readiness score between 0 and 5? \vspace{5pt}
        
        {\textbf{ Response:}}
        \end{flushleft}
    \end{minipage}
};
\node[fancytitle, rounded corners, right=10pt] at (box.north west) {Prompt template for few-shot learning};
\end{tikzpicture}%
\end{tabular}

\vspace{10pt}

\begin{tabular}{c} 
\begin{tikzpicture}
\node [mybox] (box){%
    \begin{minipage}{0.935\textwidth}
    \footnotesize
        \begin{flushleft} 
        {\textbf{ Instruction:}} You are a medical decision maker skilled in making decisions based on summarized reports. \vspace{5pt}

        {\textbf{\{\{few\_shot\_examples\}\}}} \vspace{5pt}
        
        {\textbf{ Question:}} The user is 38-year-old female with 179 cm. The analysis of recent 2-weeks sensor readings show:
        
        \begin{itemize}
            \item Steps: [5260.0, 1323.0, ..., 3308.0] steps 
            \item Burned Calories: [349.0, 97.0, ..., 206.0] calories
            \item Resting Heart Rate: [65.61, 64.33, ..., 68.26] beats/min
            \item Sleep Minutes: [406.0, 533.0, ..., 462.0] minutes
            \item Mood: 3 out of 5
        \end{itemize}
        
        In this regard, can you predict user's readiness score between 0 and 5 and provide step-by-step reasoning? \vspace{5pt}

        {\textbf{\{\{\{Prediction\}-\{Reasoning\} from LLMs\}\}}} \vspace{5pt}
        
        {\textbf{ Response:}}
        \end{flushleft}
    \end{minipage}
};
\node[fancytitle, rounded corners, right=10pt] at (box.north west) {Prompt template for few-shot CoT-SC Approach};
\end{tikzpicture}%
\end{tabular}

\end{figure*}

\begin{figure*}[t!]
\caption{Decoding methods used to generate prompts on health prediction questions.}
\centering
\begin{tabular}{c} 
\begin{tikzpicture}
\node [mybox] (box){%
    \begin{minipage}{0.935\textwidth}
    \footnotesize
        \begin{flushleft} 
        {\textbf{greedy\_search =}} 
        \{ \\
        \ \ \ \ ``num\_beams" : 1, \\
        \ \ \ \ ``do\_sample" : False, \\
        \ \ \ \ ``max\_new\_tokens" : 128, \\ 
        \ \ \ \ ``early\_stopping" : False \\
        \} \\ 
        {\textbf{beam\_serach =}} 
        \{ \\
        \ \ \ \ ``um\_beams" : 4, \\
        \ \ \ \ ``do\_sample" : False, \\
        \ \ \ \ ``max\_new\_tokens" : 128, \\ 
        \ \ \ \ ``early\_stopping" : True, \\
        \} \\ 
        {\textbf{sampling\_top\_k =}} 
        \{ \\
        \ \ \ \ ``do\_sample" : True, \\
        \ \ \ \ ``num\_beams": 1, \\
        \ \ \ \ ``max\_new\_tokens": 128, \\ 
        \ \ \ \ ``early\_stopping": True, \\
        \ \ \ \ ``temperature": 0.7, \\
        \ \ \ \ ``top\_k": 50 \\
        \} \\
        {\textbf{sampling\_top\_p =}} 
        \{ \\
        \ \ \ \ ``do\_sample" : True, \\
        \ \ \ \ ``top\_k": 0, \\ 
        \ \ \ \ ``num\_beams": 1, \\
        \ \ \ \ ``max\_new\_tokens": 128, \\ 
        \ \ \ \ ``early\_stopping": True, \\
        \ \ \ \ ``temperature": 0.7, \\
        \ \ \ \ ``top\_p": 0.9 \\
        \} \\
        {\textbf{sampling =}} 
        \{ \\
        \ \ \ \ ``do\_sample" : True, \\
        \ \ \ \ ``top\_k": 50, \\
        \ \ \ \ ``num\_beams": 1, \\
        \ \ \ \ ``max\_new\_tokens": 128,  \\
        \ \ \ \ ``early\_stopping": True, \\
        \ \ \ \ ``temperature": 0.4, \\
        \ \ \ \ ``top\_p": 0.9 \\
        \} \\
        
        \end{flushleft}
    \end{minipage}
    \label{fig:decoding}
};
\node[fancytitle, rounded corners, right=10pt] at (box.north west) {Decoding Methods for MedAlpaca};
\end{tikzpicture}%
\end{tabular}
\end{figure*}

\begin{figure*}[t!]
\caption{Model configuration for fine-tuning MedAlpaca.}
\centering
\begin{tabular}{c} 
\begin{tikzpicture}
\node [mybox] (box){%
    \begin{minipage}{0.935\textwidth}
    \footnotesize
        \begin{flushleft} 
        \{ \\ 
        \ \ \ \ ``\_name\_or\_path": "decapoda-research/llama-7b-hf", \\
        \ \ \ \ ``architectures": [ \\
        \ \ \ \ \ \ \ \ ``LlamaForCausalLM" \\
        \ \ \ \ ], \\
        \ \ \ \ ``attention\_bias": false, \\
        \ \ \ \ ``bos\_token\_id": 0, \\
        \ \ \ \ ``eos\_token\_id": 1, \\
        \ \ \ \ ``hidden\_act": ``silu", \\
        \ \ \ \ ``hidden\_size": 4096, \\
        \ \ \ \ ``initializer\_range": 0.02, \\
        \ \ \ \ ``intermediate\_size": 11008, \\
        \ \ \ \ ``max\_position\_embeddings": 2048, \\
        \ \ \ \ ``max\_sequence\_length": 2048, \\
        \ \ \ \ ``model\_type": ``llama", \\
        \ \ \ \ ``num\_attention\_heads": 32, \\
        \ \ \ \ ``num\_hidden\_layers": 32, \\
        \ \ \ \ ``num\_key\_value\_heads": 32, \\
        \ \ \ \ ``pad\_token\_id": -1, \\
        \ \ \ \ ``pretraining\_tp": 1, \\
        \ \ \ \ ``rms\_norm\_eps": 1e-06, \\
        \ \ \ \ ``rope\_scaling": null, \\
        \ \ \ \ ``rope\_theta": 10000.0, \\
        \ \ \ \ ``tie\_word\_embeddings": false, \\
        \ \ \ \ ``torch\_dtype": ``float32", \\
        \ \ \ \ ``transformers\_version": ``4.35.0.dev0", \\
        \ \ \ \ ``use\_cache": false, \\
        \ \ \ \ ``vocab\_size": 32000 \\
        \}

        \end{flushleft}
    \end{minipage}
    \label{fig:config}
};
\node[fancytitle, rounded corners, right=10pt] at (box.north west) {model configuration for training};
\end{tikzpicture}%
\end{tabular}
\end{figure*}

\begin{table*}[!htbp]
\centering
\tiny
\caption{Pairwise t-test results for baseline models vs. HealthAlpaca series models for \textbf{Task 1: STRS}. * indicates $p < 0.05$ and ** indicates $p < 0.001$.}
\begin{tabular}{c l c c c c}
\hline
\textbf{Category}  & \textbf{Model Comparison} & \textbf{HealthAlpaca-7b-lora} & \textbf{HealthAlpaca-7b} & \textbf{HealthAlpaca-13b-lora} & \textbf{HealthAlpaca-13b} \\
\hline
 \multirow{12}{*}{\rotatebox{90}{Zero-shot}} & MedAlpaca & ** & * & * & ** \\
 & PMC-Llama & ** & ** & ** & ** \\
 & Asclepius & ** & ** &  & ** \\
 & ClinicalCamel & ** &  &  &  \\
 & Flan-T5 & ** & ** &  & ** \\
 & Palmyra-Med & * & ** & * & ** \\
 & Llama-2 & * &  &  &  \\
 & BioMedGPT & ** & ** &  & ** \\
 & BioMistral & ** &  &  & * \\
 & GPT-3.5 & ** & ** & ** & ** \\
 & GPT-4 & ** & ** & ** & ** \\
 & Gemini-Pro & ** & ** & * & ** \\

\midrule

\multirow{12}{*}{\rotatebox{90}{Few-shot}} & MedAlpaca & ** & * & * & ** \\
& + CoT & * & ** & * & ** \\
& + CoT-SC & ** &  &  &  \\
& GPT-3.5 &  & ** & ** & ** \\
& + CoT & ** & * & * & ** \\
& + CoT-SC & ** &  &  & * \\
& GPT-4 & ** & * & * & ** \\
& + CoT & ** &  &  & * \\
& + CoT-SC & ** & * &  &  \\
& Gemini-Pro & ** & ** & ** & ** \\
& + CoT &  & ** & ** & ** \\
& + CoT-SC & ** & ** &  & ** \\

\midrule

\multirow{5}{*}{\rotatebox{90}{Baseline}} & MLP & * & ** & ** & ** \\
& SVM & ** & ** &  & ** \\
& RandomForest & ** & ** &  & ** \\
& BERT & * &  &  &  \\
& BioMed-RoBERTa & ** & ** &  & ** \\

\hline
\end{tabular}
\label{app:ttest1}
\end{table*}

\begin{table*}[!htbp]
\centering
\tiny
\caption{Pairwise t-test results for baseline models vs. HealthAlpaca series models for \textbf{Task 2: READ}. * indicates $p < 0.05$ and ** indicates $p < 0.001$.}
\begin{tabular}{c l c c c c}
\hline
\textbf{Category}  & \textbf{Model Comparison} & \textbf{HealthAlpaca-7b-lora} & \textbf{HealthAlpaca-7b} & \textbf{HealthAlpaca-13b-lora} & \textbf{HealthAlpaca-13b} \\
\hline
 \multirow{12}{*}{\rotatebox{90}{Zero-shot}} & MedAlpaca &  & ** & ** & ** \\
 & PMC-Llama & * & ** & ** & ** \\
 & Asclepius & * &  &  &  \\
 & ClinicalCamel &  & ** & ** & ** \\
 & Flan-T5 &  & ** & ** & ** \\
 & Palmyra-Med &  & ** & ** & ** \\
 & Llama-2 & * & ** & ** & ** \\
 & BioMedGPT & * & ** & ** & ** \\
 & BioMistral &  & ** & ** & ** \\
 & GPT-3.5 & * & ** & ** & ** \\
 & GPT-4 &  & ** & ** & ** \\
 & Gemini-Pro & ** & * & * &  \\

\midrule

\multirow{12}{*}{\rotatebox{90}{Few-shot}} & MedAlpaca &  & * & * & * \\
& + CoT & ** & ** & ** & ** \\
& + CoT-SC &  & ** & ** & ** \\
& GPT-3.5 & ** &  &  &  \\
& + CoT & * & * &  &  \\
& + CoT-SC & * &  &  &  \\
& GPT-4 &  &  &  &  \\
& + CoT &  & * & * &  \\
& + CoT-SC & ** &  &  & * \\
& Gemini-Pro &  & ** & ** & ** \\
& + CoT &  & ** & ** & ** \\
& + CoT-SC & ** &  &  & * \\

\midrule

\multirow{5}{*}{\rotatebox{90}{Baseline}} & MLP & ** & ** & ** & ** \\
& SVM & ** &  &  & ** \\
& RandomForest & ** & * &  & ** \\
& BERT & * & ** & ** & ** \\
& BioMed-RoBERTa & ** &  &  & ** \\

\hline
\end{tabular}
\end{table*}

\begin{table*}[!htbp]
\centering
\tiny
\caption{Pairwise t-test results for baseline models vs. HealthAlpaca series models for \textbf{Task 3: FATG}. * indicates $p < 0.05$ and ** indicates $p < 0.001$.}
\begin{tabular}{c l c c c c}
\hline
\textbf{Category}  & \textbf{Model Comparison} & \textbf{HealthAlpaca-7b-lora} & \textbf{HealthAlpaca-7b} & \textbf{HealthAlpaca-13b-lora} & \textbf{HealthAlpaca-13b} \\
\hline
 \multirow{12}{*}{\rotatebox{90}{Zero-shot}} & MedAlpaca &  &  & * &  \\
 & PMC-Llama & ** & ** & ** & ** \\
 & Asclepius &  &  & ** & * \\
 & ClinicalCamel & ** &  & ** &  \\
 & Flan-T5 & ** &  & ** &  \\
 & Palmyra-Med &  &  & * &  \\
 & Llama-2 &  &  & * &  \\
 & BioMedGPT & ** &  & * &  \\
 & BioMistral & ** &  & ** &  \\
 & GPT-3.5 & ** &  &  &  \\
 & GPT-4 & ** & * &  & * \\
 & Gemini-Pro &  &  & ** &  \\

\midrule

\multirow{12}{*}{\rotatebox{90}{Few-shot}} & MedAlpaca &  &  & ** &  \\
 & + CoT & ** & ** & ** & ** \\
 & + CoT-SC &  & * & ** & ** \\
 & GPT-3.5 & * &  &  &  \\
 & + CoT & ** & * &  & * \\
 & + CoT-SC &  &  & * &  \\
 & GPT-4 & ** &  & * &  \\
 & + CoT & ** & * &  & * \\
 & + CoT-SC & * &  & ** &  \\
 & Gemini-Pro &  & * & ** & * \\
 & + CoT &  & * & ** & ** \\
 & + CoT-SC & ** &  & * &  \\

\midrule

\multirow{5}{*}{\rotatebox{90}{Baseline}} & MLP &  &  & ** & * \\
 & SVM & ** &  & ** &  \\
 & RandomForest & ** &  & * &  \\
 & BERT &  &  &  &  \\
 & BioMed-RoBERTa & ** &  & ** &  \\

\hline
\end{tabular}
\end{table*}

\begin{table*}[!htbp]
\centering
\tiny
\caption{Pairwise t-test results for baseline models vs. HealthAlpaca series models for \textbf{Task 4: SQ}. * indicates $p < 0.05$ and ** indicates $p < 0.001$.}
\begin{tabular}{c l c c c c}
\hline
\textbf{Category}  & \textbf{Model Comparison} & \textbf{HealthAlpaca-7b-lora} & \textbf{HealthAlpaca-7b} & \textbf{HealthAlpaca-13b-lora} & \textbf{HealthAlpaca-13b} \\
\hline
 \multirow{12}{*}{\rotatebox{90}{Zero-shot}} & MedAlpaca &  &  & ** & ** \\
 & PMC-Llama & ** & ** & ** & ** \\
 & Asclepius & * &  &  &  \\
 & ClinicalCamel & ** &  &  &  \\
 & Flan-T5 &  &  &  &  \\
 & Palmyra-Med & ** &  &  &  \\
 & Llama-2 &  &  & * &  \\
 & BioMedGPT & * &  &  &  \\
 & BioMistral & ** &  &  &  \\
 & GPT-3.5 &  & ** & ** & ** \\
 & GPT-4 &  &  & ** & * \\
 & Gemini-Pro &  &  & ** & ** \\

\midrule

\multirow{12}{*}{\rotatebox{90}{Few-shot}} & MedAlpaca &  &  & * & * \\
 & + CoT & * &  & * & * \\
 & + CoT-SC & ** &  &  &  \\
 & GPT-3.5 &  &  & ** & ** \\
 & + CoT &  &  & * &  \\
 & + CoT-SC & ** &  &  &  \\
 & GPT-4 &  &  & * &  \\
 & + CoT &  &  & ** & * \\
 & + CoT-SC & ** &  &  &  \\
 & Gemini-Pro &  &  & ** & ** \\
 & + CoT &  & * & ** & ** \\
 & + CoT-SC & ** & ** & * & * \\

\midrule

\multirow{5}{*}{\rotatebox{90}{Baseline}} & MLP & * & ** & ** & ** \\
 & SVM & ** &  &  &  \\
 & RandomForest & * &  & * &  \\
 & BERT &  &  &  &  \\
 & BioMed-RoBERTa & ** &  &  &  \\
 
\hline
\end{tabular}
\end{table*}

\begin{table*}[!htbp]
\centering
\tiny
\caption{Pairwise t-test results for baseline models vs. HealthAlpaca series models for \textbf{Task 5: SR}. * indicates $p < 0.05$ and ** indicates $p < 0.001$.}
\begin{tabular}{c l c c c c}
\hline
\textbf{Category}  & \textbf{Model Comparison} & \textbf{HealthAlpaca-7b-lora} & \textbf{HealthAlpaca-7b} & \textbf{HealthAlpaca-13b-lora} & \textbf{HealthAlpaca-13b} \\
\hline
 \multirow{12}{*}{\rotatebox{90}{Zero-shot}} & MedAlpaca &  & ** & ** & ** \\
 & PMC-Llama &  &  &  &  \\
 & Asclepius & ** & ** & ** & ** \\
 & ClinicalCamel &  & ** & ** & ** \\
 & Flan-T5 & ** & ** & ** & ** \\
 & Palmyra-Med & * & ** & ** & ** \\
 & Llama-2 & ** & ** & ** & ** \\
 & BioMedGPT & * &  &  &  \\
 & BioMistral &  & * & * & * \\
 & GPT-3.5 &  & * & * & * \\
 & GPT-4 &  & ** & ** & ** \\
 & Gemini-Pro &  & * & * & * \\

\midrule

\multirow{12}{*}{\rotatebox{90}{Few-shot}} & MedAlpaca &  &  &  &  \\
 & + CoT &  & * & * &  \\
 & + CoT-SC &  & * & * &  \\
 & GPT-3.5 & ** & * & * & ** \\
 & + CoT & ** &  &  &  \\
 & + CoT-SC & ** & ** & ** & ** \\
 & GPT-4 & ** & * & * & ** \\
 & + CoT & ** &  &  &  \\
 & + CoT-SC & ** & * & * & ** \\
 & Gemini-Pro &  & ** & ** & ** \\
 & + CoT &  & ** & ** & ** \\
 & + CoT-SC & ** & * & * & ** \\

\midrule

\multirow{5}{*}{\rotatebox{90}{Baseline}}& MLP &  & ** & ** & ** \\
 & SVM & ** &  &  & ** \\
 & RandomForest & ** & * & * & ** \\
 & BERT & ** & ** & ** & ** \\
 & BioMed-RoBERTa & ** & ** & ** & ** \\
 
\hline
\end{tabular}
\end{table*}

\begin{table*}[!htbp]
\centering
\tiny
\caption{Pairwise t-test results for baseline models vs. HealthAlpaca series models for \textbf{Task 6: SD}. * indicates $p < 0.05$ and ** indicates $p < 0.001$.}
\begin{tabular}{c l c c c c}
\hline
\textbf{Category}  & \textbf{Model Comparison} & \textbf{HealthAlpaca-7b-lora} & \textbf{HealthAlpaca-7b} & \textbf{HealthAlpaca-13b-lora} & \textbf{HealthAlpaca-13b} \\
\hline
 \multirow{12}{*}{\rotatebox{90}{Zero-shot}} & MedAlpaca & ** & ** & ** & ** \\
 & PMC-Llama & ** & * & * & ** \\
 & Asclepius & ** & ** & ** & ** \\
 & ClinicalCamel & ** & ** & * &  \\
 & Flan-T5 & ** &  &  & ** \\
 & Palmyra-Med & ** & ** & ** & ** \\
 & Llama-2 & ** & ** & ** & ** \\
 & BioMedGPT & ** & ** & ** & ** \\
 & BioMistral & ** & ** & * &  \\
 & GPT-3.5 & ** & ** & ** & ** \\
 & GPT-4 & ** & ** & ** & ** \\
 & Gemini-Pro & * & ** &  &  \\

\midrule

\multirow{12}{*}{\rotatebox{90}{Few-shot}} & MedAlpaca & * &  &  & ** \\
 & + CoT & ** & * & * & ** \\
 & + CoT-SC & ** & ** & ** & ** \\
 & GPT-3.5 & ** &  &  & ** \\
 & + CoT & ** & * & * & ** \\
 & + CoT-SC &  & * &  &  \\
 & GPT-4 &  & * &  & ** \\
 & + CoT &  &  &  &  \\
 & + CoT-SC &  & ** &  & * \\
 & Gemini-Pro &  & * &  & ** \\
 & + CoT &  &  &  & * \\
 & + CoT-SC & ** & ** & * &  \\

\midrule

\multirow{5}{*}{\rotatebox{90}{Baseline}} & MLP & ** & ** & ** & ** \\
 & SVM & ** & ** & ** & ** \\
 & RandomForest & ** & ** & ** & ** \\
 & BERT & ** & * & * & ** \\
 & BioMed-RoBERTa & ** & * & * & ** \\
 
\hline
\end{tabular}
\end{table*}

\begin{table*}[!htbp]
\centering
\tiny
\caption{Pairwise t-test results for baseline models vs. HealthAlpaca series models for \textbf{Task 7: ANX}. * indicates $p < 0.05$ and ** indicates $p < 0.001$.}
\begin{tabular}{c l c c c c}
\hline
\textbf{Category}  & \textbf{Model Comparison} & \textbf{HealthAlpaca-7b-lora} & \textbf{HealthAlpaca-7b} & \textbf{HealthAlpaca-13b-lora} & \textbf{HealthAlpaca-13b} \\
\hline
 \multirow{12}{*}{\rotatebox{90}{Zero-shot}} & MedAlpaca &  & * & * &  \\
 & PMC-Llama & * & ** & ** & * \\
 & Asclepius &  &  & * &  \\
 & ClinicalCamel &  & ** & ** &  \\
 & Flan-T5 & ** & ** & ** & ** \\
 & Palmyra-Med & ** & ** & ** & ** \\
 & Llama-2 &  & ** & ** &  \\
 & BioMedGPT &  & * & * &  \\
 & BioMistral &  & ** & ** &  \\
 & GPT-3.5 & ** & ** & ** & ** \\
 & GPT-4 & ** & ** & ** & ** \\
 & Gemini-Pro &  & ** & ** &  \\

\midrule

\multirow{12}{*}{\rotatebox{90}{Few-shot}} & MedAlpaca &  & * & * &  \\
 & + CoT &  & * & * &  \\
 & + CoT-SC &  & ** & ** & * \\
 & GPT-3.5 & ** & ** & ** & ** \\
 & + CoT &  & ** & ** &  \\
 & + CoT-SC &  &  &  &  \\
 & GPT-4 &  & * & * &  \\
 & + CoT &  & ** & ** &  \\
 & + CoT-SC & ** & ** &  & ** \\
 & Gemini-Pro &  & ** & ** & * \\
 & + CoT &  & ** & ** &  \\
 & + CoT-SC &  & * & * &  \\

\midrule

\multirow{5}{*}{\rotatebox{90}{Baseline}} & MLP &  &  &  &  \\
 & SVM & ** & ** & * & ** \\
 & RandomForest &  & ** & ** & * \\
 & BERT & ** & ** & ** & ** \\
 & BioMed-RoBERTa & ** & ** & ** & ** \\
 
\hline
\end{tabular}
\end{table*}

\begin{table*}[!htbp]
\centering
\tiny
\caption{Pairwise t-test results for baseline models vs. HealthAlpaca series models for \textbf{Task 8: DEP}. * indicates $p < 0.05$ and ** indicates $p < 0.001$.}
\begin{tabular}{c l c c c c}
\hline
\textbf{Category}  & \textbf{Model Comparison} & \textbf{HealthAlpaca-7b-lora} & \textbf{HealthAlpaca-7b} & \textbf{HealthAlpaca-13b-lora} & \textbf{HealthAlpaca-13b} \\
\hline
 \multirow{12}{*}{\rotatebox{90}{Zero-shot}} & MedAlpaca &  & ** & * &  \\
 & PMC-Llama & ** & ** & ** & ** \\
 & Asclepius & ** & ** & ** & ** \\
 & ClinicalCamel &  & * &  &  \\
 & Flan-T5 & ** & ** & ** & ** \\
 & Palmyra-Med & ** & ** & ** & ** \\
 & Llama-2 &  & * & * & * \\
 & BioMedGPT &  & * & * &  \\
 & BioMistral & ** & ** & ** & ** \\
 & GPT-3.5 & ** & ** & ** & ** \\
 & GPT-4 & ** & ** & ** & ** \\
 & Gemini-Pro &  & ** & * & * \\

\midrule

\multirow{12}{*}{\rotatebox{90}{Few-shot}} & MedAlpaca & ** &  &  &  \\
 & + CoT &  & * & * & * \\
 & + CoT-SC & ** &  &  &  \\
 & GPT-3.5 &  &  &  &  \\
 & + CoT & ** &  &  &  \\
 & + CoT-SC & ** &  &  &  \\
 & GPT-4 & ** &  &  &  \\
 & + CoT & * &  &  &  \\
 & + CoT-SC & ** &  &  & * \\
 & Gemini-Pro &  & ** & * & ** \\
 & + CoT &  & * & * &  \\
 & + CoT-SC & ** & ** &  & ** \\

\midrule

\multirow{5}{*}{\rotatebox{90}{Baseline}} & MLP &  & * & * &  \\
 & SVM & ** & * &  &  \\
 & RandomForest & ** & ** & * &  \\
 & BERT & ** & ** & ** & ** \\
 & BioMed-RoBERTa & ** & ** & ** & ** \\
 
\hline
\end{tabular}
\end{table*}

\begin{table*}[!htbp]
\centering
\tiny
\caption{Pairwise t-test results for baseline models vs. HealthAlpaca series models for \textbf{Task 9: ACT}. * indicates $p < 0.05$ and ** indicates $p < 0.001$.}
\begin{tabular}{c l c c c c}
\hline
\textbf{Category}  & \textbf{Model Comparison} & \textbf{HealthAlpaca-7b-lora} & \textbf{HealthAlpaca-7b} & \textbf{HealthAlpaca-13b-lora} & \textbf{HealthAlpaca-13b} \\
\hline
 \multirow{12}{*}{\rotatebox{90}{Zero-shot}} & MedAlpaca &  &  & * & * \\
 & PMC-Llama & ** & ** & ** & ** \\
 & Asclepius & ** & ** & ** & ** \\
 & ClinicalCamel & ** & * & * & ** \\
 & Flan-T5 & ** &  & * & ** \\
 & Palmyra-Med & ** &  &  &  \\
 & Llama-2 & ** & ** & ** & ** \\
 & BioMedGPT & * & * & * & ** \\
 & BioMistral &  & * & * & ** \\
 & GPT-3.5 &  & * & * & * \\
 & GPT-4 &  &  & * & * \\
 & Gemini-Pro & ** & * & * & ** \\

\midrule

\multirow{12}{*}{\rotatebox{90}{Few-shot}} & MedAlpaca &  &  & * &  \\
 & + CoT & * & * & * & ** \\
 & + CoT-SC &  & * & * & ** \\
 & GPT-3.5 &  &  &  &  \\
 & + CoT & ** &  & * & ** \\
 & + CoT-SC & ** &  &  &  \\
 & GPT-4 & * & * & * & ** \\
 & + CoT &  &  & * &  \\
 & + CoT-SC & * &  & * & ** \\
 & Gemini-Pro &  & * & * & ** \\
 & + CoT & * & * & * & ** \\
 & + CoT-SC & ** & ** &  & ** \\

\midrule

\multirow{5}{*}{\rotatebox{90}{Baseline}} & MLP &  & * & * & ** \\
 & SVM & ** & * & * & ** \\
 & RandomForest & ** &  &  & * \\
 & BERT & ** & ** &  & ** \\
 & BioMed-RoBERTa & ** & ** &  & ** \\
 
\hline
\end{tabular}
\end{table*}

\begin{table*}[!htbp]
\centering
\tiny
\caption{Pairwise t-test results for baseline models vs. HealthAlpaca series models for \textbf{Task 10: CAL}. * indicates $p < 0.05$ and ** indicates $p < 0.001$.}
\begin{tabular}{c l c c c c}
\hline
\textbf{Category}  & \textbf{Model Comparison} & \textbf{HealthAlpaca-7b-lora} & \textbf{HealthAlpaca-7b} & \textbf{HealthAlpaca-13b-lora} & \textbf{HealthAlpaca-13b} \\
\hline
 \multirow{12}{*}{\rotatebox{90}{Zero-shot}} & MedAlpaca &  &  &  &  \\
 & PMC-Llama &  &  & * & * \\
 & Asclepius &  &  &  &  \\
 & ClinicalCamel &  &  &  &  \\
 & Flan-T5 & ** & ** & ** & ** \\
 & Palmyra-Med & ** & ** & ** & ** \\
 & Llama-2 & ** & ** & ** & ** \\
 & BioMedGPT & ** & ** & ** & ** \\
 & BioMistral &  &  &  &  \\
 & GPT-3.5 &  & ** &  & * \\
 & GPT-4 & ** & ** & ** & ** \\
 & Gemini-Pro &  & ** &  & ** \\

\midrule

\multirow{12}{*}{\rotatebox{90}{Few-shot}} & MedAlpaca &  &  &  &  \\
 & + CoT &  & ** &  & * \\
 & + CoT-SC &  & ** &  & ** \\
 & GPT-3.5 &  & ** &  & * \\
 & + CoT &  & ** &  &  \\
 & + CoT-SC &  & ** &  &  \\
 & GPT-4 &  & ** &  & ** \\
 & + CoT &  & ** &  & * \\
 & + CoT-SC &  & ** &  & ** \\
 & Gemini-Pro &  & ** &  &  \\
 & + CoT &  &  &  &  \\
 & + CoT-SC &  & ** &  & * \\

\midrule

\multirow{5}{*}{\rotatebox{90}{Baseline}} & MLP &  &  & * & ** \\
 & SVM &  & ** &  & ** \\
 & RandomForest &  & ** &  & ** \\
 & BERT &  & ** & ** & ** \\
 & BioMed-RoBERTa &  & ** & * & ** \\
 
\hline
\end{tabular}
\label{app:ttest10}
\end{table*}

\begin{table*}[t!]
\tiny
\caption{\textbf{Consumer Health Tasks}. We curate ten tasks with four datasets, across four domains. For the Dataset Size, we indicate the number of QA samples and the label distribution in the parentheses.}
\begin{tabular}{>{\rotatebox{90}}p{1.3cm} >{\centering\arraybackslash}p{1.2cm} >{\centering\arraybackslash}p{3.0cm} p{1.4cm} p{5.4cm} p{1.5cm}}
\toprule
\centering\textbf{Domain} & \centering\textbf{Dataset} & \centering\textbf{Task} & \centering\textbf{Metric} & \centering\textbf{Dataset Size} & \textbf{\makecell[l]{Text Length \\ (Token)}} \\
\midrule
\multirow{6}{*}{MHealth} & \multirow{1}{*}{PMData} & \multirow{1}{*}{Stress Prediction} & \centering \multirow{1}{*}{MAE} \ $\downarrow$ & 1784 (0: 1, 1: 21, 2: 315, 3: 833, 4: 240, 5: 8)  & $202 \pm 42$ \\ \\
& \multirow{1}{*}{LifeSnaps} & \multirow{1}{*}{Stress Resilence Prediction} & \centering \multirow{1}{*}{MAE} \ $\downarrow$& \multirow{1}{*}{150 (0-1: 30, 1-2: 67, 2-3: 35, 3-4: 13, 4-5: 5)} & $202 \pm 42$ \\ \\ 
& \multirow{2}{*}{GLOBEM} & \multirow{1}{*}{\makecell{Estimate of PHQ4 Depression \\ Estimate of PHQ4 Anxiety}} & \centering \multirow{2}{*}{MAE} \ $\downarrow$ & \multirow{2}{*}{\makecell{1735 (0: 748, 1: 555, 2: 296, 3: 108, 4: 28) \\1735 (0: 1093, 1: 372, 2: 194, 3: 63, 4: 13)}} & \multirow{2}{*}{$216 \pm 0$} \\ \\ 
\midrule   
\multirow{4}{*}{Activity} & \multirow{2}{*}{PMData} & \multirow{1}{*}{\makecell{Readiness Prediction \\ Fatigue Prediction}} & \centering \multirow{2}{*}{\makecell{{\multirow{1}{*}{MAE} \ $\downarrow$} \\ {\centering \multirow{1}{*}{Accuracy} \ $\uparrow$}}}  & \multirow{1}{*}{\makecell[tl]{1420 (0-1: 32, 2-3: 217, 4-5:628, 6-7: 423, 8-10: 120) \\ 1418 (1: 53, 2: 410, 3: 845, 4: 101, 5: 9)}} & \multirow{2}{*}{$202 \pm 42$} \\
\end{tabular}
\end{table*}

\begin{table*}[!t]
\centering
\tiny
\caption{\textbf{Performance Evaluation of LLMs on Health Prediction Tasks.} \textbf{STRS}: Stress, \textbf{READ}: Readiness, \textbf{FATG}: Fatigue, \textbf{SQ}: Sleep Quality, \textbf{SR}: Stress Resilience, \textbf{SD}: Sleep Disorder, \textbf{ANX}: Anxiety, \textbf{DEP}: Depression, \textbf{ACT}: Activity, \textbf{CAL}: Calories. ``-'' denotes the failure cases due to token size limit or unreasonable responses. ``N/A'' denotes the case where the prediction is not reported or cannot be conducted. For each column (task), the best result is \textcolor{darkgreen}{\textbf{bolded}}, and the second best is \underline{underlined}. CoT denotes the chain-of-thoughts and SC denotes the self-consistency prompting. For each task, arrows in the parenthesis indicate the desired direction of improvement. $\uparrow$ indicates higher values are better for macro F1-score, while $\downarrow$ indicates lower values are better for mean absolute percentage error (MAPE).}
\setlength{\tabcolsep}{4.0pt}
\begin{tabular}{p{0.1cm} p{2.37cm} p{1.085cm} p{1.085cm} p{1.085cm} p{1.085cm} p{1.085cm} p{1.085cm} p{1.085cm} p{1.085cm} p{1.085cm} p{1.085cm} p{1.085cm} p{1.085cm} p{1.085cm}}
\toprule
\multicolumn{1}{c}{} & \multicolumn{1}{c}{\textbf{Dataset}} & \multicolumn{4}{c}{\textbf{PMData}} & \multicolumn{2}{c}{\textbf{LifeSnaps}} & \multicolumn{2}{c}{\textbf{GLOBEM}} & \multicolumn{2}{c}{\textbf{AW\_FB}} \\
\cmidrule(lr){3-6} \cmidrule(lr){7-8} \cmidrule(lr){9-10} \cmidrule(lr){11-12} 
 \textbf{} & \textbf{Model} & \textbf{\makecell{STRS \\ ($\downarrow$)}} & \textbf{\makecell{READ \\ ($\downarrow$)}} & \textbf{\makecell{FATG \\ ($\uparrow$)}} & \textbf{\makecell{SQ \\ ($\downarrow$)}} & \textbf{\makecell{SR \\ ($\downarrow$)}} & \textbf{\makecell{SD \\ ($\uparrow$)}} & \textbf{\makecell{ANX \\ ($\downarrow$)}} & \textbf{\makecell{DEP \\ ($\downarrow$)}} & \textbf{\makecell{ACT \\ ($\uparrow$)}} & \textbf{\makecell{CAL \\ ($\downarrow$)}} \\
\midrule

\multirow{12}{*}{\rotatebox{90}{Zero-shot}} & MedAlpaca & 54.0 \fontsize{4.5pt}{4.5pt}\selectfont{\textpm \ 0.0} & 49.2 \fontsize{4.5pt}{4.5pt}\selectfont{\textpm \ 0.0} & 0.26 \fontsize{4.5pt}{4.5pt}\selectfont{\textpm \ 0.0} & 38.3 \fontsize{4.5pt}{4.5pt}\selectfont{\textpm \ 0.0} & 87.1 \fontsize{4.5pt}{4.5pt}\selectfont{\textpm \ 0.0} & 0.22 \fontsize{4.5pt}{4.5pt}\selectfont{\textpm \ 0.0} & 66.7  \fontsize{4.5pt}{4.5pt}\selectfont{\textpm \ 0.0} & 72.2 \fontsize{4.5pt}{4.5pt}\selectfont{\textpm \ 0.0} & 0.13 \fontsize{4.5pt}{4.5pt}\selectfont{\textpm \ 0.0} & 77.1 \fontsize{4.5pt}{4.5pt}\selectfont{\textpm \ 0.0}   \\
\end{tabular}
\end{table*}

\end{document}